%% file: 0_main.tex
\documentclass[sigconf]{acmart}

\usepackage{makecell} 
\usepackage{pifont} 
\usepackage{algorithm}
\usepackage{algpseudocode}

\definecolor{labelyellow}{RGB}{255,200,0}

\newcommand{\yellowcircle}[2][labelyellow]{%
  \begingroup
  \makebox[1.05em][c]{%
    \ooalign{%
      \hfil
      \raisebox{-0.58ex}{%
        \textcolor{#1}{%
          \fontsize{15}{15}\selectfont\ding{108}%
        }%
      }%
      \hfil\cr
      \hfil
      \raisebox{0.02ex}{%
        \textcolor{black}{%
          \sffamily\bfseries\normalsize #2%
        }%
      }%
      \hfil\cr
    }%
  }%
  \endgroup
}

\AtBeginDocument{%
  }

\setcopyright{acmlicensed}

\newcommand{\eg}[0]{e.g.,}
\newcommand{\ie}[0]{i.e.,}

\newif\ifshowcomments

\showcommentsfalse  
\ifshowcomments
    \newcommand{\bl}[0]{\color{blue}}
    \newcommand{\david}[1]{\color{red}\textbf{David comment:} #1\color{black}}
    \newcommand{\dedit}[1]{\color{orange}#1\color{black}}
    \newcommand{\puqi}[1]{}
    \newcommand{\help}[1]{\color{green!75!black}\textbf{Need help:} #1\color{black}}
\else
    \newcommand{\bl}{\color{black}}
    \newcommand{\david}[1]{}
    \newcommand{\dedit}[1]{}
    \newcommand{\puqi}[1]{}
    \newcommand{\help}[1]{}
\fi

\newcommand{\quotes}[1]{\emph{``#1''}}

\newcommand{\system}{\textit{Attune}}

\definecolor{Author1}{HTML}{377eb8}

\copyrightyear{2026}
\acmYear{2026}
\setcopyright{cc}
\setcctype{by}
\acmConference[UIST '26]{The 39th Annual ACM Symposium on User Interface Software and Technology}{November 02--05, 2026}{Detroit, MI, USA}
\acmBooktitle{The 39th Annual ACM Symposium on User Interface Software and Technology (UIST '26), November 02--05, 2026, Detroit, MI, USA}
\acmDOI{10.1145/3830398.3830514}
\acmISBN{979-8-4007-2856-3/2026/11}

\begin{document}

\author{Puqi Zhou}
\orcid{0000-0002-6486-8883}
\affiliation{%
  \institution{Department of Computer Science}
  \institution{George Mason University}
  \city{Fairfax}
  \state{VA}
  \country{USA}
}
\email{pzhou@gmu.edu}

\author{Sungsoo Ray Hong}
\orcid{0000-0001-6050-5404}
\affiliation{%
  \institution{Department of Information Sciences and Technology}
  \institution{George Mason University}
  \city{Fairfax}
  \state{VA}
  \country{USA}
}
\email{shong31@gmu.edu}

\author{David Porfirio}
\orcid{0000-0001-5383-3266}
\affiliation{%
  \institution{Department of Computer Science}
  \institution{George Mason University}
  \city{Fairfax}
  \state{VA}
  \country{USA}
}
\email{dporfiri@gmu.edu}

\title{\system{}: A Self-Annotation Tool for Understanding Robot Operator Attention Profiles}

\begin{abstract}

Deploying robot fleets in complex, real-world environments requires human operators to supervise multiple robots simultaneously. Managing operator attention is a fundamental challenge of designing multi-robot supervision interfaces, encompassing both feed layout and feed content (\ie{} robot \textit{behavior} design). 
Thus far, designers lack empirical guidance on the latter---how to change a robot's behavior to capture, sustain, or relinquish operator attention during multi-robot supervision. In our vision of the future, designers should be able to use this guidance to calibrate robot behavior to different operator attention profiles.
Treating operator eye gaze as a robot behavior design clue, we created a 
{\bl pre-deployment elicitation tool} called \system{}. 
\system{} automatically identifies \textit{when} meaningful gaze shifts occur, provides AI assistance for annotating \textit{why} shifts occurred, and outputs a summary of operator gaze patterns for 
{\bl operator review}.
We evaluated \system{} through a user study in which participants annotated the visual triggers that drew their attention.
Our findings unveil variation in {\bl observed} gaze patterns and reveal how \system{} helps characterize operator attention.
\end{abstract}



\begin{teaserfigure}
  \includegraphics[width=\textwidth]{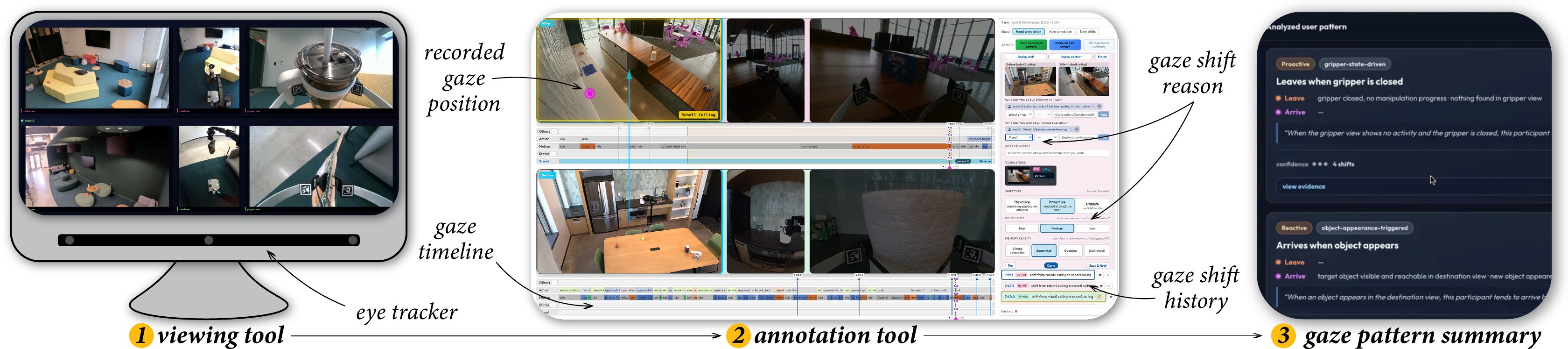}
  \caption{\system{} enables robot operators to self-annotate attentional shifts and functions as follows: \yellowcircle{1} an operator watches robots while \system{} records eye gaze; \yellowcircle{2} the operator plays back their gaze shifts and retroactively attributes causes to these shifts; and \yellowcircle{3} \system{} co-annotates additional shifts and extracts patterns to be presented to the operator for confirmation.}
  \label{fig:teaser}
\end{teaserfigure}

\maketitle
\input{1_introduction}
\input{2_relatedwork}
\input{3_systemdesign}
\input{4_evaluation}
\input{5_discussion}
\section{Conclusion}
We present \system{}, a {\bl pre-deployment elicitation} tool that helps operators construct attention profiles for the future calibration of multi-robot monitoring interfaces and robot behavior. \system{} records operator eye gaze when watching robots perform different tasks and assists the operator in attributing possible causes to their gaze shifts. \system{} then uses individual shifts to extrapolate the operator's gaze patterns. We conducted a user study to evaluate \system{} and uncovered several design implications.

\begin{acks}
This work was supported by George Mason University and in part by 4-VA, a higher education collaborative partnership for advancing the Commonwealth of Virginia.
\end{acks}

\bibliographystyle{ACM-Reference-Format}
\bibliography{cite}

\appendix

\end{document}

%% file: 1_introduction.tex
\section{Introduction}

Multi-robot fleets are increasingly common on college campuses \cite{kaya2023design} and hospitals \cite{diana2024designing} for making deliveries, in warehouses for stocking and sorting items \cite{zhang2025new}, cities and rural areas for searching for missing people \cite{wang2025concurrent}, in addition to many other environments. While each robot in the fleet is mostly autonomous in practice, robot \textit{operators} are still needed to supervise and manage each robot and to intervene or troubleshoot when necessary. Often, a single operator is tasked with monitoring multiple robots and must maintain attention to key events and intervene in a robot's operation. Many interfaces that facilitate this monitoring consist of multiple adjacent robot camera feeds streamed to a single screen  \cite{zhou2026designing, gao2014grid}.

The design of such interfaces is crucial to maintaining operator performance, which prior work indicates is affected by operator stress, workload, and attention \cite{sam2024impact, srivastava2014attention}.
Attention, in particular, has led to efforts to improve operator interface design \cite{crandall2010computing}, although this is difficult because operators possess individual differences due to personal psychology, task-related attentional biases, or being influenced by the setting within which the operator resides \cite{wickens2021attention, wolfe2007guided}. Multi-robot viewing interfaces should therefore be customized for individual operator attention needs.

The first step towards customizing a multi-robot viewing interface for a specific operator is to understand the operator's attention profile. 
{\bl For this paper, we refer to an \textit{attention profile} as an operator-reviewed collection of recurring gaze-shift patterns, grounded in the operator’s retrospective rationales and the robot and environmental context surrounding those shifts.}
\puqi{We will define “Attention profile” in \textbf{§1}/§3 as an operator-confirmed account of when/why gaze shifts occur and how robot motions/environmental features influence attention.}
In order to assess the operator's individual attention profile, including what interface features and robot actions they are most prone to looking at or being distracted by, we envision a \textit{calibration step} that occurs when multi-robot fleets are deployed in a particular setting. Using this information, a designer can create or customize an interface tailored specifically to the operator or tweak the robots' behaviors in order to appropriately allocate operator attention and mitigate distractions.
{\bl In this paper, we take a first step towards the latter, characterizing how robot behavior coincides with gaze shifts and how calibration can structure these shifts into attention profiles.}
\puqi{As R1 proposed, we will revise Abstract/\textbf{§1}/§3/§5 to frame Attune as a pre-deployment elicitation tool that surfaces and structures operators' retroactive accounts of gaze shifts into “attention profiles,” leaving downstream uses to future work.}

We therefore constructed \system{}, a {\bl pre-deployment elicitation tool} that facilitates this calibration step {\bl by helping characterize cross-robot gaze shifts on minimal, two-robot configurations, leaving larger configurations to future work}. Shown in Figure \ref{fig:teaser}, \system{} first uses eye tracking to record operator gaze during a {\bl two}-robot video observation step, selects \textit{valid} gaze shifts, and then presents the operators with an \textit{annotation tool} that guides them in {\bl retroactively} attribute possible causes to a subset of these shifts. \system{} then uses large language models (LLMs) to further annotate valid shifts and produce a summary of the operator's attention profile.

{\bl Our evaluation of \system{} uses the two-robot setup as a foundational first step for validating the general feasibility of our proposed workflow, including cross-robot gaze shift production and subsequent annotation. Specifically, we conducted a user study that asks} (\textbf{RQ1}) how should annotation tools be designed to elicit retroactive causal attributions for gaze shifts and (\textbf{RQ2}) what are the characteristics of gaze shifts during multi-robot supervision.
\puqi{tone down large-fleet claims and frame the two-robot setup as a foundational first step for validating general feasibility, and the minimal configuration that produces the cross-robot gaze shifts necessary to evaluate Attune’s elicitation workflow.}
Our results highlight the benefits of structured replay and contextual scaffolding and uncover both shared and variable gaze patterns between participants and scenarios. 
Our contributions are:

\begin{itemize}

    \item {\bl\textit{System}---\system{}, a tool that elicits gaze-shift attributions.}
    \puqi{Addressing 1AC’s limited technical contribution concern, we will clarify that Attune primarily presents a system contribution: integrating multiple components into a novel interaction paradigm that makes tacit gaze-shift rationales inspectable.}
    \item \textit{Empirical}---an evaluation of \system{}'s ability to capture gaze shift attributions, operator gaze patterns, and usability.
    \item \textit{Design}---design implications for improving self-annotation tools for achieving our vision of operator gaze calibration.

\end{itemize}
\color{black}

%% file: 2_relatedwork.tex
\section{Related Work}
This work draws from prior research on human attention, interfaces for robot operators, and interfaces for data annotation.
%
%
\subsection{Human Attention}
The \textit{Premotor Theory of Attention} ties the preparation of eye motor movement to attentional changes, thus linking gaze shifts (\textit{saccades}) to attention \cite{rizzolatti1987reorienting}. In the context of human-robot supervisory control, there are several factors that predict attentional shifts towards different \textit{areas of interest (AOIs)}---the \textit{salience} of the AOI (\eg{} its visual distinctiveness), the \textit{expectancy} of observing events at a particular AOI, the \textit{value} of looking at the AOI, and the \textit{effort} of the shift \cite{wickens2017attentional, wickens2015noticing, wickens2021attention}. Crucially, however, eye tracking alone cannot determine the cause of a shift. Furthermore, attention shifts can be a ``top-down'' (\eg{} knowledge-driven) process \cite{wickens2021attention}, and can thus differ from person to person depending on an individual's own goals and experiences \cite{wolfe2007guided}. These limitations provide a motivation for using \system{} to rationalize, or attribute possible causes to, gaze shifts.
There is a substantial amount of prior work on both predicting and leveraging attention to manage human-centered computing~\cite{wu2024theia}. Recently, \citet{pei2025attentionar} created \textit{AttentionAR}, an augmented-reality interface that monitors user attention and strategically overlays hazard alerts in the user's headset. Managing human attention is also highly relevant in human-robot interaction, where robot behavioral design is critical to managing operator attention \cite{ozsu2025distraction}. Prior work shows how to measure \cite{lemaignan2016real}, interpret \cite{schirmer2025utilizing}, guide \cite{miyashita2025framelight, rea2017movers}, and respond to \cite{yamaoka2009developing} human attention shifts. As further justification for our own work, \citet{saad2019welcoming} show that robot behaviors can be chosen to appropriately manage human attention.
\subsection{Interfaces for Robot Operators}
%
\textit{Robot operators} encompass a broad range of skilled professionals whose responsibilities depend on contextual expectations and the autonomous capabilities of the robot being operated, ranging from direct control (\ie{} teleoperation) \cite{pelikan2025people, elbeleidy2023beyond, gamboa2025we, walker2024cyber} to monitoring semi-autonomous systems with only minimal intervention needed \cite{benford2025charting}. Interventions are often physical, such as clearing obstacles \cite{pelikan2025making} and preventing bystander harm \cite{benford2025tangles}. This paper instead focuses on \textit{remote} monitoring \cite{lee2025minding, ebadi2023present}, which often occurs during the deployment of semi-autonomous ``sidewalk'' delivery robots \cite{kaya2023design, valdez2023humans, grimm2021practicalities}, search-and-rescue robots \cite{ahlskog2024fostering}, and robots that traverse areas that are hazardous to humans \cite{chiou2021mixed}.{\bl ~In mature deployments, operators are immersed in camera feeds for extended
windows under either continuous monitoring ~\cite{chiou2022towards, zhou2026designing,
foroughi2023near} or manage-by-exception alert response paradigms ~\cite{lewis2014task, chien2011effects, gao2014grid}. \system{} does not assume a fixed monitoring duration after calibration, intending to apply to both sustained monitoring and brief (\eg{} alert-triggered) windows.}
\puqi{For real-world applicability, we will cite prior deployments in §2.2 where operators are immersed in continuous feeds [Chiou et al.IJSR2022, Zhou et al.CHI2026]
We will clarify that Attune does not assume the monitoring duration after the calibration step; its elicited insights are intended to apply to both longer and shorter bursts of sustained monitoring (e.g., immediately post-alert).
}

Interfaces must be carefully designed for these operators. In human-robot interaction, operator performance is uniquely affected by how the robot presents its autonomy, the reliability of its sensors, and the duration of its behaviors, among other aspects \cite{drury2007adapting, drury2007modeling}. \citet{roy2023need} find that how the robot presents its autonomy to a remote operator affects situation awareness. Poorly designed interfaces that fail to manage these factors can lead to high operator workload, poor memory retention, misallocated attention, and, in general, reduced operator effectiveness \cite{peters2015human}. This has led to the careful development of remote viewing interfaces such as \citet{zhou2026designing}'s multi-video ground robot sensemaking interface.
\begin{figure*}
 \includegraphics[width=\textwidth]{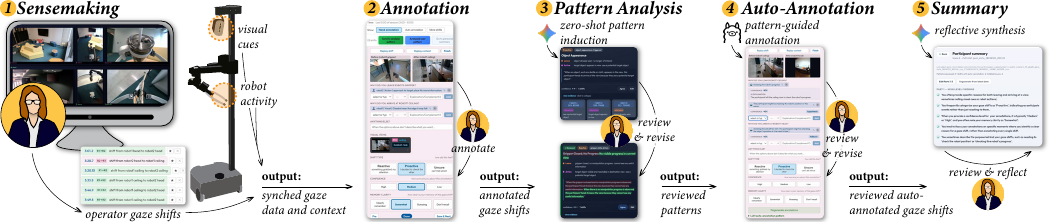}
 \caption{\bl{The \system{} pipeline, with the output being a synthesized attention profile summarized for the operator to review.}}
 \label{fig:system pipline}
\end{figure*}
\subsection{Behavioral Annotation Interfaces}
Behavioral annotation draws from a long lineage of interfaces for qualitative analysis, much of which focuses on analyzing text content \cite{atlasti, dhakal2022nvivo, lewis2007qda}. Our work instead draws from multimodal (\eg{} video and audio) human \textit{behavioral} annotation. \textit{BORIS} \cite{friard2016boris} is one example, which enables annotators to assign qualitative codes to video and audio content. \textit{DOTE} \cite{mcilvenny2022dote, mcilvenny2024dotebase, mcilvenny2024guide} and \textit{ELAN} \cite{sloetjes2008annotation, elan} serve a similar purpose, and are popular in the human-centered computing community for conducting conversational (and more generally \textit{interactional}) analyses between humans or between humans and robots \cite{pelikan2024designing, passero2024honkable, pelikan2024encountering}. Beyond \textit{ROSAnnotator}, few such tools exist specifically for human-robot interaction \cite{zhang2025rosannotator}. None discussed thus far are intended for annotating human eye gaze behavior.

Instead, we require an interface that (1) tracks operator eye gaze behavior; (2) automatically identifies significant gaze shifts (or \textit{saccades}) versus stationary gaze fixations; (3) assists the operator in attributing possible causes to a subset of these saccades; and (4) automatically attributes causes to previously unseen saccades. Prior work satisfies subsets of these requirements. Existing research already addresses the binary classification between fixations and saccades \cite{vortmann2021imaging, iddrisu2026eye}. Similarly, classifying \textit{what} a viewer fixates on, though laborious without assistance from eye gaze trackers \cite{somashekarappa2020annotation}, is usually an automatic process \cite{deane2023deep}. The \textit{eyeNotate} tool, for example, semi-automatically classifies \textit{areas of interest} from a viewer's gaze \textit{fixation events} \cite{barz2025eyenotate}. In addition to mapping fixation events to real-world targets, \textit{gazeMapper}~\cite{niehorster2025gazeMapper} facilitates analysis of gaze behavior in multi-party settings. 
With all of these aforementioned tools, however, the causes of saccades are unknown. We specifically seek a solution that enables viewers themselves (the \textit{operators}) to rationalize both top-down and salience-driven saccades.

%% file: 3_systemdesign.tex
\section{System Design}\label{sec:system}
%

We designed \system{} as a {\bl pre-deployment} operator calibration facilitator for multi-robot monitoring, focusing on {\bl a minimal (two-robot) variant}. The system combines multi-video monitoring, eye-tracking-based gaze recording, and both manual and LLM-assisted annotation in order to capture where operators look, and crucially, help them explain \textit{why} they shifted their attention. Through the five phases shown in Figure~\ref{fig:system pipline}, we aim to 
{\bl surface} individual operator attention profiles, or their {\bl self-confirmed accounts of recurring gaze shift rationales}.
These phases form a one-time, human-in-the-loop, pre-deployment calibration step: (1) video sensemaking to capture operator gaze; (2) gaze-shift annotation; (3) cross-shift pattern analysis; (4) pattern-guided auto-annotation generation; and (5) synthesis and summarization of the operator's attention profile. 

\textbf{Scenario.} We use a scenario to assist in our description of \system{}. Consider a company that is deploying a fleet of robots at a hospital. The robots make deliveries between staff and patients, while a \textit{robot operator} observes the fleet remotely and intervenes when necessary. It is crucial that the multi-robot supervision interface used by the operator effectively manages operator attention. For example, if one robot in the fleet is performing a high-risk task, the others should not draw the operator's attention.
A designer who works for the company is responsible for ensuring that the behaviors of each robot in the fleet are appropriately calibrated to the operator's attention profile and uses \system{} as a calibration facilitator. 
%
%
\subsection{Phase 1: Multi-Robot Video Sensemaking}
%
\begin{figure*}[!t]
    \centering
    \includegraphics[width=\textwidth]{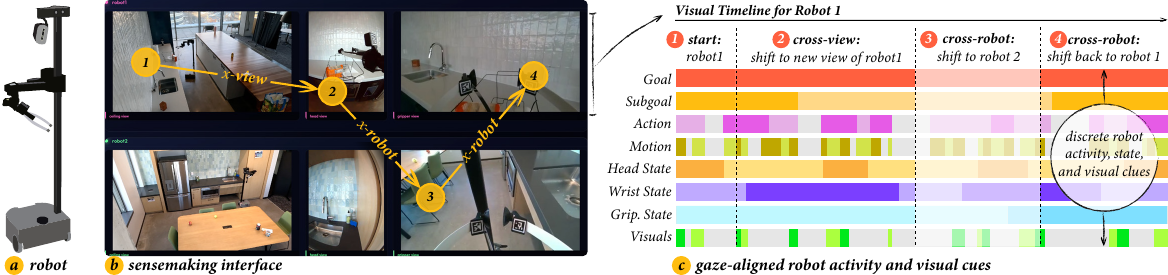}
    \caption{The sensemaking interface shows two \yellowcircle{a} robots conducting tasks simultaneously, in which both robots stream \yellowcircle{b} three video feeds to the interface. The interface tracks operator gaze, {\bl including cross-robot (x-robot) and cross-view (x-view) gaze shifts}, which is aligned with \yellowcircle{c} robot activity and visual cues.}
    \label{fig:sensemaking}
\end{figure*}

Figure \ref{fig:sensemaking} depicts our sensemaking interface, which supports viewing two robots concurrently executing independent tasks. Each robot is accompanied by three synchronized camera views from the \textit{ceiling}, the robot's \textit{head} point-of-view (POV), and its \textit{gripper} POV, ordered from left to right. In total, this yields six synchronized video streams. This design is intended to facilitate the characterization of the operator's gaze shifts both within the same robot (\textit{cross-view} shifts) and across different robots (\textit{cross-robot} shifts).

While the operator monitors the robot feeds, \system{} records their eye gaze. Gaze is saved both as \textit{pixel-level} gaze records (\eg{} the exact position of eye gaze on the monitor) and \textit{discretized} gaze records that specify which video region is being viewed at each moment. Gaze is synchronized with video playback to enable direct correspondence between gaze behavior and video events.

\small
\begin{table}[b]
    \centering
    \caption{Our hierarchical definition of robot activity.}
    \label{tab:robot}
    \begin{tabular}{p{0.19\columnwidth}p{0.74\columnwidth}} 
        \textbf{Rob. Activity} & \textbf{Labels} \\
        \hline
        \makecell{Goal} & \makecell[l]{Domain specific. Examples include \textit{tidy kitchen}.}\\
        \addlinespace[4px]
        \makecell{Subgoal} & \makecell[l]{Domain specific. Examples include \textit{wash dishes}.}\\
        \makecell{\\Action} & \textit{Open, close, grab, carry, drop, navigate, approach, transport, wait, notify, activate, deactivate, pull, push, adjust [limb], change view, remain stationary.}\\
        \makecell{\\Motion} & \textit{Gripper open/close, base movement, arm retract/extend, arm lift/down, head move, wrist rotate, idle.} \\
        \addlinespace[4px]
        \makecell{Gripper State} & \makecell[l]{Discrete based on openness: \textit{closed, partially open, open.}} \\
        \addlinespace[-10px]
        \makecell{\\Wrist State} & Pitch, yaw, and roll discretized into combined labels: \textit{neutral, up, down, left, forward, right, (strong) rolled, stowed.}\\
        \makecell{\\Head State} & Combinations of pan and tilt: \textit{(far) left, forward, (far) right, back, (strongly) down, level, (strongly) up. }\\
    \end{tabular}
\end{table}

\normalsize

In addition to video time, gaze is aligned with discrete \textit{robot activity}, inspired by the hierarchical abstractions that are common in the task planning literature \cite{holler2020hddl, fox2003pddl2}. Shown in Table \ref{tab:robot}, the different levels of robot activity include (a) the high-level goal currently being pursued by the robot; (b) the subgoal that the robot is currently executing in pursuit of the larger goal; (c) the current action being executed; (d) the robot's low-level motion; and (e) the robot's joint state. In practice, this information can come from the robot itself. 

In addition, \system{} uses YOLO \cite{redmon2016you} to extract and segment visual cues from each video, which are stored as object labels and maskable visual regions. As a result, each gaze shift can be interpreted with reference not only to video time, but also to ongoing, discrete robot behaviors and visible task-relevant objects at that time.

\textbf{Sensemaking in Action:} In our scenario, the operator sits in front of a computer monitor and watches both robots conduct different tasks. Shown in Figure \ref{fig:sensemaking}, \system{} records and maps their gaze to robot activity and visual cues.

\subsection{Phase 2: Data Annotation Interface}
\begin{figure*}[!t]
    \centering
    \includegraphics[width=\textwidth]{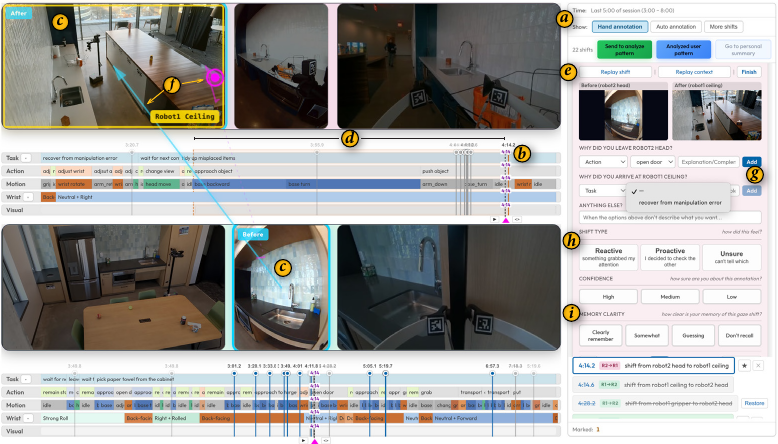}
    \caption{The annotation tool. The annotation card \yellowcircle{a} enables operators to annotate specific shifts \yellowcircle{b} that occur between different robot views \yellowcircle{c}. Operators can control replay during annotation (\yellowcircle{d} and \yellowcircle{e}), view their {\bl recent gaze shift (the blue arrow) and gaze point (the pink dot)} \yellowcircle{f}, attribute a cause for the shift (\yellowcircle{g} and \yellowcircle{h}) and their confidence \yellowcircle{i}.}
    \label{fig:annot}
\end{figure*}

After the sensemaking session, the operator enters the \textit{annotation tool} shown in Figure~\ref{fig:annot}, a dedicated workspace for retroactively rationalizing gaze shifts. {\bl This phase is designed specifically to occur after (\ie{} rather than alongside) the sensemaking phase to avoid disrupting natural operator gaze patterns.}
The annotation tool merges valid gaze shifts from Phase~1---cross-robot and cross-view---into a time-sorted list. For each robot, the operator can inspect time-aligned tracks beneath the videos that show contextual information: goal, subgoal, action, motion, joint state, and visual cues.

The first part of annotation is automatic \textit{gaze shift selection}, followed by operator \textit{shift inspection and replay}, which is facilitated by \textit{nonlinear timeline magnification}. The operator then engages in \textit{structured annotation} to retroactively attribute causes to gaze shifts. Note that structured annotation is subject to memory degradation.

{\bl
\begin{algorithm}[t]
\bl
\caption{Selecting Gaze Shifts for Review}
\label{alg:gaze-shift-sampling}
\begin{algorithmic}[1]
\State Split video $V$ into temporally equal bins $v \in \mathcal{V}$
\State $\mathcal{P}, \mathcal{M}, \mathcal{A} \gets [\ ]$
\Comment{Curated, manual, and auto pools}
\ForAll{$v \in \mathcal{V}$}
    \State \textbf{skip} $v$ if no valid shift exists for either type
    \Repeat
        \State $g_r, g_v \gets \Call{SampleShifts}{v}$
        \Comment{cross-robot/view}
    \Until{$\Call{Curate}{g_r} \wedge \Call{Curate}{g_v}$} \Comment{See Table~\ref{tab:gaze-shift-rules}} 
    \State $\Call{Append}{\mathcal{P}, (v, g_r, g_v)}$
\EndFor
\State $\mathcal{P} \gets \Call{Shuffle}{\mathcal{P}}$
\State Put $m$ shifts of each type in $\mathcal{M}$ \Comment{Manual pool}
\State Put $n$ shifts of each type in $\mathcal{A}$ \Comment{Auto pool}
\end{algorithmic}
\end{algorithm}
}

\textbf{Gaze Shift Selection:}\label{sec:gaze-shifts}
\system{} automatically {\bl selects and curates} gaze shifts for the user to annotate {\bl from the set of observed gaze shifts. In order to capture the operator's stable viewing rhythm while reducing early-session novelty effects and recall decay,} selection and curation only occur within the most recently viewed segments, specifically the final five minutes of each session. Algorithm \ref{alg:gaze-shift-sampling} depicts the selection procedure. \puqi{NOTE I EXPLAIN THIS in here 3.2 instead of section 4: On 2AC’s final-5-minute sampling clarification, we will clarify in §4 that this design choice aims to avoid novelty effects, capture a stable viewing rhythm, and mitigate retroactive recall decay via recent viewings. }
\system{} first partitions the videos into temporally equal bins (Line 1). \system{} then samples (Line 6) and attempts to curate (Line 7) a cross-robot and cross-view gaze shift from within each bin, where curation is successful if any of the five rules in Table \ref{tab:gaze-shift-rules} apply for both shifts. Curated gaze shifts include shifts from one camera view (A) to another (B) where gaze dwells on both the source view A and destination view B for at least 100ms (a typical minimum gaze fixation duration based on \citet{salvucci2000identifying}). Gaze episodes shorter than 100ms on either view are excluded as saccadic pass-throughs or transitional movements.
Sometimes, gaze shifts originate from or target portions of the interface that are outside of the six camera views. We designate these as \textit{bridge shifts} to or from an unknown region. 
Bridge shifts from A$\rightarrow$ unknown or unknown$\rightarrow$ B are successfully curated if the time spent in the unknown region is at least 200ms ({\bl an operational} threshold {\bl informed} by \citet{manor2003defining}). 
Most shifts are sampled randomly to preserve ecological validity; a small subset is quality-anchored by selecting shifts with the longest dwell-before durations, ensuring the inclusion of high-confidence, unambiguous shifts alongside temporally representative ones.

\system{} divides curated shifts into a set $\mathcal{M}$ for manual annotation (Line 11) and $\mathcal{A}$ for auto-annotation (Line 12, see Phase~4).
Each set contains an equal number of cross-robot and cross-view shifts.  \system{} overlays gaze-shift indicators on the timeline to help operators quickly locate moments of attention transition.


\small
\begin{table}[t]
\bl
\centering
\caption{Rules for  curating valid gaze shifts. \textit{A}, \textbf{B}, and \textit{C} are different robot views (cross-view or cross-robot). \textit{Unknown} denotes areas of the screen that do not correspond to a view.}
\label{tab:gaze-shift-rules}
\small
\setlength{\tabcolsep}{4pt}
\begin{tabular}{@{}p{0.76\columnwidth}l@{}}
\toprule
\textbf{Raw shift} &
\textbf{Curated shift} \\
\midrule
$A$ ($\geq 100$\,ms) $\rightarrow$ $B$ ($\geq 100$\,ms)
& $A \rightarrow B$ \\[2pt]
$A$ ($\geq 100$\,ms) $\rightarrow$ $B$ ($< 100$\,ms) $\rightarrow$ $C$ ($\geq 100$\,ms)
& $A \rightarrow C$ \\[2pt]
$A$ ($\geq 100$\,ms) $\rightarrow$ \textit{unknown} ($< 200$\,ms) $\rightarrow$ $B$ ($\geq 100$\,ms)
& $A \rightarrow B$ \\[2pt]
\textit{unknown} ($\geq 200$\,ms) $\rightarrow$ $B$ ($\geq 100$\,ms)
& \textit{unknown} $\rightarrow B$ \\[2pt]
$A$ ($\geq 100$\,ms) $\rightarrow$ \textit{unknown} ($\geq 200$\,ms)
& $A \rightarrow$ \textit{unknown} \\
\bottomrule
\end{tabular}
\end{table}
\normalsize

\textbf{Shift Inspection and Replay:} 
\system{} displays $\mathcal{M}$ on the timeline and in a list at the bottom right side of the screen from which operators can select curated gaze shifts. Selecting a shift opens an annotation card in the right panel (Figure \ref{fig:annot} \yellowcircle{a}).
\system{} then presents the shift timestamp (Figure \ref{fig:annot} \yellowcircle{b}), the \emph{from} and \emph{to}
camera views (Figure \ref{fig:annot} \yellowcircle{c}), and the temporally aligned robot context surrounding the shift (Figure \ref{fig:annot} \yellowcircle{d}). To support recall, the system offers two replay modes: (1) a \emph{replay shift } that focuses on the transition between \emph{from} and \emph{to} views, and 
(2) a \emph{replay context} that begins 5s
before the shift and ends 0.5s after (Figure \ref{fig:annot} \yellowcircle{e}). During both replays, the
\emph{from} and \emph{to} views receive directional
overlays and the gaze point and recent trajectory are drawn on screen (Figure \ref{fig:annot} \yellowcircle{f}). The remaining four views are dimmed.

\textbf{Nonlinear Timeline Magnification:}
To further assist with causal attribution, the system applies a piecewise
nonlinear temporal mapping over the interval $[t-90\text{s},\,
t+30\text{s}]$. The earliest segment $[t-90, t-30]$ occupies 30\% of
the timeline width; the immediately preceding segment $[t-30, t]$
occupies 60\% and is highlighted with an orange overlay; and the
post-shift segment $[t, t+30]$ occupies the remaining 10\%. \system{} thus magnifies the 30s most likely to contain the causal event while
preserving earlier history and post-shift context in compressed form
for a quick overview.

\textbf{Structured Annotation:}
Within an annotation card, the operator provides structured
explanations for why they left the previous view (\emph{why leave}) and why they arrived at the new view (\emph{why arrive}), retroactively drawing causal links between their gaze shift and the contextual information that may have drawn their shift (Figure \ref{fig:annot} \yellowcircle{g}). \system{} also offers a free-response text box if none of these action or visual categories apply. Operators also label the \textit{shift type}, or whether their gaze shift was driven by ``bottom-up'' (labeled \textit{reactive} gaze shifts by the interface, reflecting involuntary shifts) or ``top-down'' (labeled \textit{proactive} gaze shifts by the interface, intended to reflect voluntary gaze shifts) mechanisms (Figure \ref{fig:annot} \yellowcircle{h}). Operators additionally provide confidence of their annotations and a rating of their memory clarity (Figure \ref{fig:annot} \yellowcircle{i}). Overall, the annotation card organizes operator reasoning and preserves the flexibility to capture nuanced or ambiguous motivations in free text. 

\begin{figure}[!t]
    \centering
    \includegraphics[width=\linewidth]{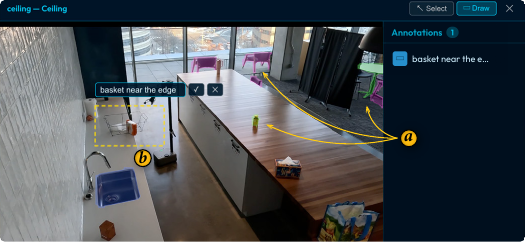}
    \caption{\system{} highlights YOLO-recognized objects \yellowcircle{a} and enables operators to label unclassified regions \yellowcircle{b}.}
    \label{fig:visual}
\end{figure}

For attributing non-robot visual cues to gaze shifts, operators can select YOLO-segmented objects from the relevant video
frame, re-label objects if necessary (\eg{} if misclassified), and draw masks over unclassified areas of the view to select and label objects not caught by YOLO. Operators can then attach the resulting cue to the annotation card.
Figure \ref{fig:visual} depicts visual annotation.

\textbf{Annotation in Action:} 
In our scenario, after the monitoring session, \system{} guides the operator to review a curated set of their gaze shifts selected from the final few minutes of recording. 
From the sample, the operator selects a shift in which they moved their gaze from one robot's gripper camera to another robot's head camera, and uses the replay tool to re-examine the transition. After watching the segment, the operator explains in the annotation card that they left the gripper view because the manipulation had completed, and arrived at the head view because the second robot appeared to be approaching a person.
\system{} records this structured explanation alongside any robot and visual context at the shift.

\subsection{Phase 3: Attention Profile Pattern Analysis}
\begin{figure}[t]
    \centering
    \includegraphics[width=\linewidth]{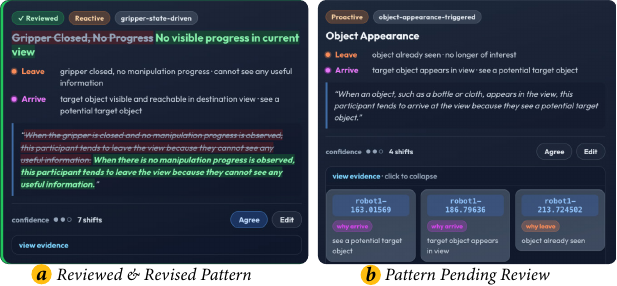}
    \caption{A pattern reviewed and corrected by the operator \yellowcircle{a}, and a pattern that is pending operator review \yellowcircle{b}.}
    \label{fig:pattern}
\end{figure}

We define a \emph{pattern} as an operator-specific, recurring
explanatory structure for multiple gaze shifts---for example, leaving views when manipulation actions complete, or arriving at new views when task-relevant objects become visible.
Figure~\ref{fig:pattern} depicts our pattern analysis interface. Once an operator has annotated a set of gaze shifts, the system supports abstraction from individual cases to higher-level viewing patterns. 

\textbf{Pattern Inference:}
The \system{} backend organizes the annotated shifts into a structured payload {\bl containing the why-leave and why-arrive rationales, robot activities, visual cues, gaze-timing features, and operator ratings,} and submits it to a zero-shot LLM-based clustering stage (Gemini-2.5-Flash).\footnote{The prompts and code for \system{} are available at \href{https://ari-lab-gmu.github.io/Attune_UIST26/}{\nolinkurl{https://ari-lab-gmu.github.io/Attune_UIST26/}}.} 
The LLM clusters shifts primarily by the semantic similarity of operators' \emph{why-leave} and \emph{why-arrive} explanations, 
{\bl grounds}
the resulting groups against shared robot and visual context, and produces candidate patterns accompanied by scope descriptions, supporting evidence, and LLM confidence. 
\puqi{We will disclose LLM prompt schemas and context serialization in-text and in our OSF repo (p.5)}
Memory-clarity and self-confidence ratings from Phase~2 serve as auxiliary reliability signals rather than primary clustering criteria.

\textbf{Operator Review and Refinement:}
The system presents candidate patterns to the operator for
review and refinement. For each pattern, the interface displays the LLM-generated description alongside the supporting annotation cards, allowing the operator to replay individual gaze shifts and revisit their original reasoning before accepting or modifying the pattern.
The operator may revise any patterns and confirm those that accurately represent their monitoring strategy. Approved patterns are saved to a \emph{reviewed pattern library}, which is carried forward in subsequent phases.

\textbf{Pattern Analysis in Action:}
In our scenario, \system{} infers candidate patterns from the operator's annotated shifts and presents each one with its supporting annotation cards. The operator can inspect the evidence, replay the linked gaze shifts, and verify whether the inferred description matches their original reasoning. If needed, they can revise the pattern before saving it. Figure~\ref{fig:pattern}b shows a pattern pending review, while Figure~\ref{fig:pattern}a shows a reviewed pattern and how the operator can refine the description. Confirmed patterns are added to the attention profile.

\subsection{Phase 4: Pattern-Guided Auto-Annotation}

{\bl As a continuation of the one-time, human-in-the-loop calibration, \system{} then} uses the operator's attention profile pattern to accelerate
annotation of additional gaze shifts, using LLM assistance to reduce annotation burden while engaging operators for verification. 
{\bl Specifically, \system{} propagates the {\bl operator-annotated shifts and operator-confirmed patterns} to unseen shifts selected from the auto-annotation set $\mathcal{A}$, rather than inferring them from scratch.} 
\puqi{Regarding R2/1AC’s concerns about manual annotation practicality, we will highlight Attune as a one-time, human-in-the-loop calibration step to elicit operator-endorsed rationales to support future full automation.}
For each shift that the operator selects from $\mathcal{A}$, \system{} constructs a structured prompt combining local shift context, pre- and post-shift robot state summaries, YOLO-derived visual cues, gaze timing features, and the operator's reviewed pattern library. This prompt is submitted to a fast, low-cost LLM (Groq Llama-3.3-70B-Versatile),\footnotemark[1] which proposes candidate annotations for \emph{shift type}, \emph{why leave}, and \emph{why arrive}, each accompanied by a confidence level and a brief explanation of the reasoning behind the suggestion.

The generated annotations are treated as editable suggestions rather than final labels. Within each annotation card, the operator can auto-generate or regenerate annotations at any time, inspect the proposed reasoning, and accept, modify, or discard each suggestion. The operator retains full responsibility for the final interpretation and must still provide confidence and memory-clarity judgments after reviewing the generated content. 

\textbf{Auto-Annotation in Action:} {\bl Auto-annotation is intended to be much faster than manual annotation.} In our scenario, the operator selects a new gaze shift from the auto-annotation pool. The system retrieves the shift context and matches it to the reviewed pattern library. The operator inspects the suggested reasoning, confirms, edits, or deletes the \emph{why leave} explanation, and revises the \emph{why arrive} explanation to reflect that the transition was triggered by a visible object rather than task anticipation. The amended annotation is saved alongside its confidence and memory clarity ratings. 

\subsection{Phase 5: Personal Summary and Reflection}


In the final phase, \system{} synthesizes the operator's complete annotation history (both manually and auto-annotated shifts) and the reviewed pattern library to produce \system{}'s final output: an \emph{operator-facing summary} generated with Gemini-2.5-Flash.\footnotemark[1]

The first part is a \emph{reflective annotation summary} that
characterizes the operator's annotation style---for example, how
consistently they applied structured labels, how often they relied on free-text reasoning, and the distribution of their confidence and memory-clarity ratings across shifts. This gives operators a transparent picture of how they engaged with the annotation process itself.
The second part presents the operator's \emph{attention profile summary}, describing the reviewed patterns that characterize their monitoring strategy---for example, which robot states or visual events most reliably triggered attention shifts, and whether the operator tended toward cross-robot or cross-view transitions. This gives operators a consolidated view of their own attention tendencies as captured and verified across Phases 2 and 3.
The third part shows explicitly how \system{} operationalized those patterns during auto-annotation in Phase~4---surfacing which patterns were invoked most frequently, how they mapped onto the auto-generated \emph{why leave} and \emph{why arrive} suggestions, and which suggestions the operator accepted, revised, or discarded.
Within the same part, a shift-level view further presents, for each shift, how the operator engaged with the auto-annotation, including whether they accepted, revised, or discarded the suggestions, and what those choices reveal about the coverage and limitations of the inferred pattern library. We aim to help operators understand how the system used their attention profile and where the inferred patterns may need further refinement.

\textbf{Summary and Reflection in Action:}
After completing the auto-annotation phase, \system{} presents the operator with their personal summary. The operator reviews the summary, inspects which patterns drove the auto-annotations in Phase~4, and reflects on where they accepted or revised the generated suggestions. The operator edits the summary to note any refinements, and the updated record is saved to their personal summary, completing the calibration session. This summary is then delivered to the designer. 

\subsection{Implementation}
We implemented \system{} as a desktop application\footnotemark[1] integrated with a stationary Tobii Pro Spark (60\,Hz) eye tracker. The system consists of a React/TypeScript frontend and a FastAPI backend that communicates with the eye tracker through a Tobii SDK. The frontend supports synchronized video playback, gaze replay, annotation, timeline visualization, and summary presentation. In the current version of \system{}, video playback is pre-recorded. The backend handles gaze recording, annotation persistence, LLM orchestration, and staged output storage. 
{\bl 
The choice to use two different LLMs rather than one reflects a quality-latency trade-off: Phase 3 clusters each operator's shifts once per session and Phase 5 generates a full operator summary, so they benefit from stronger semantic reasoning (Gemini-2.5-Flash), whereas Phase 4 generates per-shift suggestions at high throughput, favoring a faster, lower-cost model (Llama-3.3-70B). The choice to use LLMs rather than a vision-language model (VLM) is motivated by the visual context being already symbolized---gaze shifts are reasoned over YOLO-derived object labels and annotated robot activity rather than raw pixels. VLM integration, conditioned on operator-reviewed visual patterns, is left for future work.
}
\puqi{reply to 2AC/R1: We will explain in §3 that the model split is a quality/latency decision: Gemini-2.5-Flash for semantic clustering (Phase 3), Llama-3.3-70B for high-throughput per-shift suggestions (Phase 4).We use LLMs rather than VLMs to predict gaze-shift annotations by reasoning over metadata where visual cues arise from symbolic YOLO outputs and annotated robot activity, not raw pixels.}



%% file: 4_evaluation.tex
\section{Evaluation}\label{sec:evaluation}
We evaluated \system{} with an IRB-approved user study in which participants used \system{} to monitor robots and annotate their gaze. 

\subsection{Procedure}
Prior to the study, we constructed a video dataset
comprising three scenarios involving a physical Stretch 3 robot \cite{kemp2026stretch}---\textit{kitchen activities}, \textit{living room activities},
and \textit{hallway missing person search}. Each scenario was recorded twice from three synchronized vantage points---ceiling, robot head, and robot gripper---resulting in three pairs of synchronized video triplets, ranging from 35 to 83 minutes in length. Videos in the same pair were captured on nearby days and at similar times of day to preserve comparable lighting conditions and overall scene appearance.
By simultaneously playing videos within a pair, we mimic multi-robot supervision of two robots.

\begin{figure}[t]
    \centering
    \includegraphics[width=\linewidth]{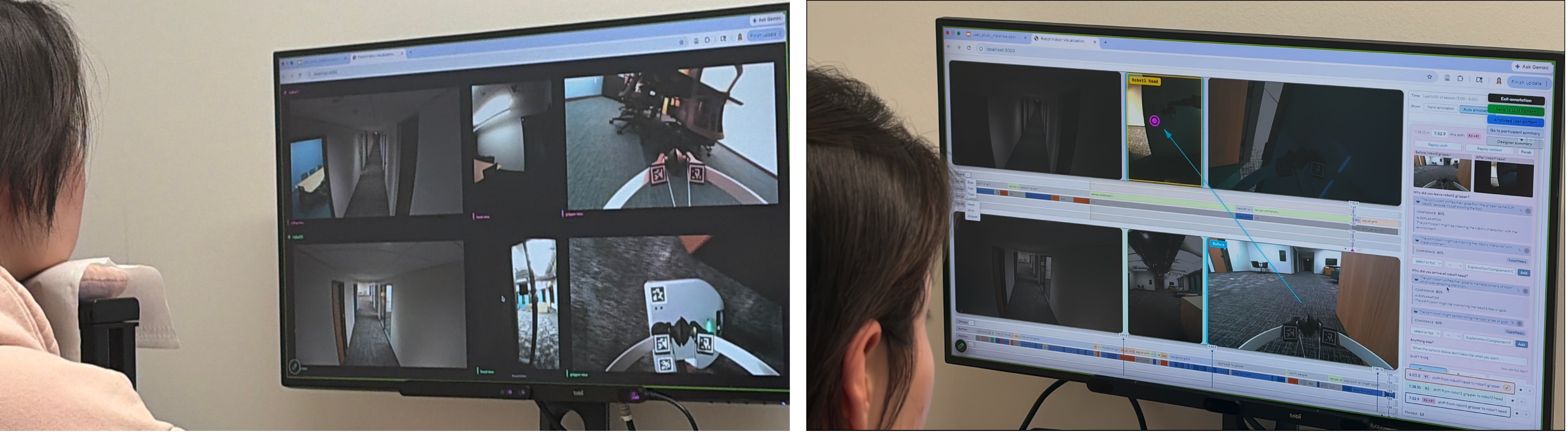}
    \vspace{-15pt}
    \caption{The setup for our evaluation.}
    \label{fig:setup}
\end{figure}

After the consent stage and eye-tracker setup, the study consisted of three phases: \textit{monitoring}, \textit{annotation}, and \textit{review}, depicted in Figure~\ref{fig:setup}. {\bl During \textit{monitoring}, each participant used \system{} to watch eight minutes of video from one scenario. We assigned four participants per scenario. For each participant, the position of each robot on the sensemaking interface (\ie{} top or bottom) and the starting timestamps for the eight-minute video duration were randomized, resulting in a unique combination of sensemaking videos per participant.} 
Our choice to have a unique combination per participant {\bl reflects a deliberate breadth-over-depth study design that reduced within-scenario statistical power but provided broader qualitative coverage and limited overfitting to any single setting.}
\puqi{On 2AC’s single-scenario concern, we will clarify in §4 that this was a deliberate breadth-over-depth exploratory choice. While reducing within-scenario power, it emphasizes rich qualitative insights and avoids overfitting } Participants were instructed to verbalize aloud to the robot in order to keep them engaged with the videos and prevent them from losing focus. Next, during \textit{annotation}, participants reviewed and annotated the rationales behind a subset of curated gaze shifts from $\mathcal{M}$ (5-8 shifts) and $\mathcal{A}$ (5 shifts), including whether the shifts were reactive or proactive, and their annotation confidence. They also rated memory clarity for each annotation. {\bl As described in \S\ref{sec:system}, the \textit{annotation} phase was designed to occur \textit{after} (\ie{} not co-occur with) the monitoring phase in order to avoid disrupting participants' natural gaze behavior.}
\puqi{we will justify post-hoc annotation in §4/§5.1 as a design choice to avoid interrupting monitoring or altering natural gaze behavior}
Finally, during \textit{review}, participants reviewed the AI-generated patterns, auto-annotated gaze shifts, and their participant-facing summary, and then answered questions about their experience.

\subsection{Participants}
We recruited 12 {\bl na\"ive} participants (8 male, 4 female) from the Mason Square campus at {\bl George Mason University}.
Four participants wore corrective lenses; all reported normal or corrected-to-normal vision, and none had prior experience with gaze-based annotation systems. Participants were compensated at \$30\,USD per hour.

\subsection{Data and Measures}
Our data consists of: (1) eye gaze recorded continuously via the eye tracker at 60\,Hz; (2) user annotations, including why-leave and why-arrive responses, shift types, confidence ratings, and memory clarity; (3) interviews via phase-embedded questions, where after each phase of the study, we asked participants a targeted question corresponding to that phase and, if time permitted, additional questions at the end of the session; (4) audio; and (5) questionnaires.

{\bl We checked the audio to characterize participants' verbalization behaviors and analyzed participant gaze patterns to assess individual differences and group-level similarities. We relied on a custom questionnaire, participant annotations, and interview data to assess \system{}'s ability to elicit causal attributions for gaze shifts. Lastly, we measured usability and usefulness via both the System Usability Scale (SUS) \cite{brooke1996sus} and interview data.}


\begin{figure*}[t!]
  \includegraphics[width=\textwidth]{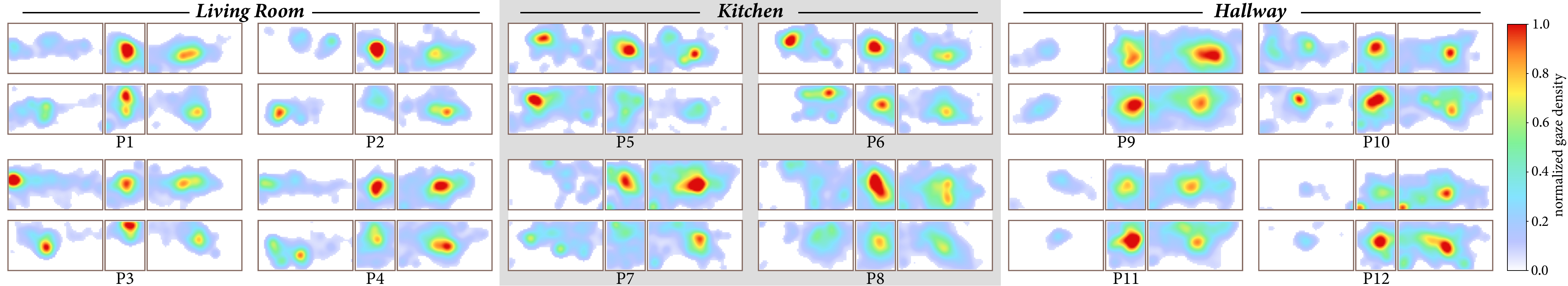}
  \caption{Participant-level gaze heatmaps across the three scenarios: Living Room, Kitchen, and Hallway. Each panel shows one participant’s normalized gaze density on the same six-view robot video interface. Warmer colors indicate higher gaze density.}
  \label{fig:heatmap}

\end{figure*}

\begin{figure}[b!]
  \includegraphics[width=\columnwidth]{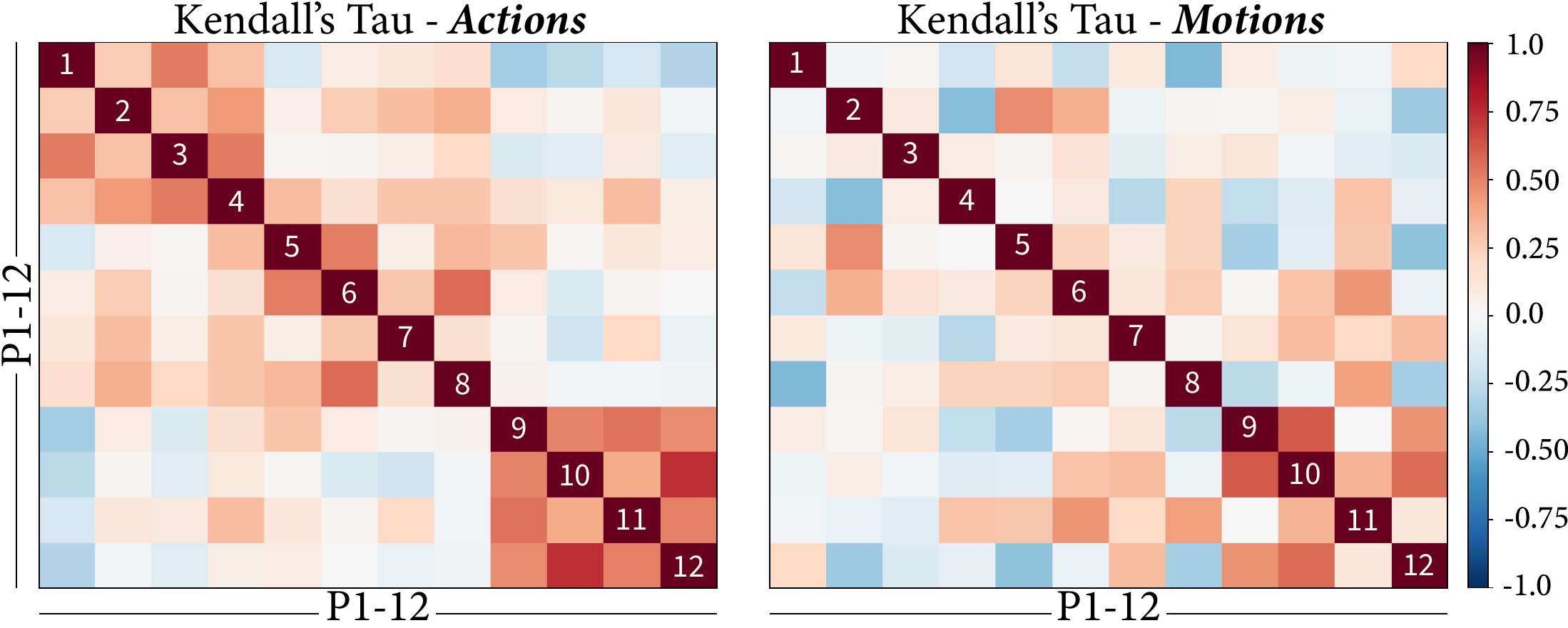}
  \caption{Cross-robot action and motion correlation matrices.}
  \label{fig:matrix}
\end{figure}

\subsection{Results}
\subsubsection{Verbalization Behaviors}
All participants verbalized aloud to comment on both robots during the monitoring session. Participants adopted different strategies: some spoke to the robots continuously as if they were companions, whereas others only verbalized when they noticed something worth warning about. 
{\bl Because speech production and eye gaze behavior occurred concurrently, these verbalization strategies may have affected participants' observed gaze patterns. It is therefore necessary to interpret participant-level differences as conditioned by our monitoring and verbalization protocol rather than as definitive evidence of natural attentional traits.}
Indeed, one participant noted that focusing on speaking made it harder to later recall gaze shift rationales.
\puqi{Regarding R1's concern that verbalization confounds visual attention/individual variation, we will acknowledge this limitation in §5.4, walk back \textbf{individual-variation claims} in §4.4.2, and sharpen our justification for verbalization as maintaining on-screen engagement. }

\subsubsection{Gaze Patterns}
{\bl The following analyses characterize observed gaze allocation, but do not adjudicate whether observed individual shifts are primarily task-driven or salience-driven, as our study does not assume a normative gaze pattern for these open-ended monitoring tasks, nor does it include independent task-based or visual-saliency benchmarks. A normative ground truth would require a separate causal model built from larger expert data.}
\puqi{R1 remarked on distinguishing salience-driven from task-driven gaze shifts, and 2AC questioned the ground-truth gaze pattern expectations. }
Figure~\ref{fig:heatmap} shows how participants distributed visual attention across \system{}'s sensemaking interface. Rather than monitoring all six views uniformly, participants appeared to concentrate on a subset of camera regions within each scenario, forming clear hotspots. 
Meanwhile, gaze patterns varied across participants, with some participants focusing attention on a few views, whereas others distributed attention more broadly.
Across the three scenarios, the spatial distribution of attention shifted as well. 


In order to quantify participant gaze patterns, we focus our analysis on how many times each participant ``shifts in'' to robot actions and motions, normalized by the prevalence of the action or motion. Specifically, we define an \textit{attraction score} as the ratio of the activity's share of valid cross-robot shift-in events to its proportional duration in the session. Scores above 1.0 indicate the behavior attracted gaze shifts more than would be expected from its occurrence frequency alone, while scores below 1.0 indicate under-attraction relative to its prevalence. 
Our subsequent analyses are cross-robot only, as cross-view transitions involved minimal changes in robot activity and showed high redundancy, whereas cross-robot shifts captured the meaningful allocation of attention across robots in response to their activities.


Our analysis of attraction scores across all participants for each motion and action category found that fine-grained manipulator motions and goal-directed actions scored consistently above baseline (score $> 1.0$), while gross locomotion and idle states attracted cross-robot shifts below baseline. {\bl This indicates} that participants were most reliably drawn to the other robot during precise manipulation and task execution and that some robot behaviors may act as shared attentional anchors. Still, participant-level monitoring strategies introduced meaningful variation in cross-robot gaze allocation, highlighting the need for personalized calibration.

We then computed pairwise rank correlation between attraction scores using Kendall’s $\tau$, and overall agreement using Kendall’s $W$ across all participants and within scenario-based subgroups (P1--P4 for \textit{living room}, P5--P8 for \textit{kitchen}, and P9--P12 for \textit{hallway}). Figure \ref{fig:matrix} depicts our results.
For actions, overall agreement was modest but significant ($W = 0.24$, $\chi^2(21) = 61.1$, $p < .001$), indicating that \textit{participants exhibited some shared patterns in which actions drew attention}. This agreement was substantially stronger within scenario groups ($W = 0.54$--$0.70$, all $p < .01$), with pairwise $\tau$ values showing \textit{more consistent gaze patterns among participants viewing the same scenario and less consistent patterns across scenarios}.
For motion, overall agreement was lower and only marginally significant ($W = 0.14$, $\chi^2(12) = 19.9$, $p = .07$), \textit{suggesting greater heterogeneity in how participants attended to motion}. However, this variability was not uniform: participants P9--P12 showed relatively strong agreement ($W = 0.57$, $p = .007$), while agreement was moderate for P5--P8 ($W = 0.40$, $p = .08$) and minimal for P1--P4 ($W = 0.18$, $p = .74$). Pairwise $\tau$ values further reveal substantial participant-dependent variation.

Taken together, these {\bl exploratory} results suggest that participants were most similar to one another in how they attended to actions within a shared scenario, while differences emerged more strongly across individuals and scenarios, particularly in how motion was prioritized, pointing to both scenario-dependent and participant-dependent variation in gaze shift patterns. {\bl However, because each participant viewed a different video pairing and participants adopted different verbalization strategies, these differences cannot be attributed solely to stable individual or scenario-based attentional tendencies. We have yet to separate out session-level variation under the present monitoring and verbalization protocol.}
\puqi{Regarding R1's concern that verbalization confounds visual attention/individual variation, we will acknowledge this limitation in §5.4, walk back \textbf{individual-variation claims} in §4.4.2, and sharpen our justification for verbalization as maintaining on-screen engagement. }

\subsubsection{{\bl Effectiveness of} Causal Attribution}
Participants generally perceived \system{} as helping them make their gaze-shift reasoning explicit. 
On a 7-point Likert scale (1 = strongly disagree, 7 = strongly agree) responding to the question of whether their personal summary is ``reflective and helpful,'' participants' average response was $M = 5.92$ ($SD = 0.67$), indicating agreement that the system captured participants' attention-allocation strategies. 

{\bl We also examined \system{}'s ability to distinguish proactive and reactive gaze.  Cross-checking proactive/reactive labels against
participants' corresponding free-response why-leave and why-arrive rationales from their annotations, there is evidence that both forms of gaze exist in the study task and that participants understood their meaning.
Proactive-labeled shifts often invoked deliberate monitoring goals
(e.g., \textit{``I just wanted to check how Robot~2 was doing,''} P11), indicating that participants did not describe all gaze shifts in terms of visual salience alone. Reactive-labeled shifts often invoked perceptually salient events (e.g., \textit{``I saw a large bottle-like object at the top of the screen, so I needed to verify,''} P2). 
Still, self-labels and free-response rationales are inherently retrospective and
cannot truly establish whether the underlying attentional
mechanism is top-down or bottom-up. We therefore interpret them
as participant-endorsed rationales rather than causal classifications.}
\puqi{We will acknowledge in §4 that our study does not check operator gaze against baseline top-down/bottom-up expectations, so it cannot objectively distinguish task-driven and salience-driven shifts. A saliency baseline will be listed as future work. However, addressing R1, we conducted a new analysis of participant “proactive/reactive” annotations, suggesting the study was not purely saliency-driven. "Proactive" shifts often matched task-driven reasons(e.g., "just want to check how robot 2 is doing"), and "Reactive" shifts often matched bottom-up (e.g., "I saw a big bottle-like object at the top of the screen, so I needed to verify”). Since self-labels and human noise alone cannot adjudicate the underlying causal mechanism, we will scope claims tightly to operator-endorsed rationales. }

Using Braun and Clarke's reflexive thematic analysis approach~\cite{braun2019thematic}, we further analyzed the interviews for \system{}'s performance as a retrospective causal attribution tool. Four themes emerged.

\textbf{Theme 1:} \textit{Annotation as a Scaffold for Externalizing Monitoring Reasoning.}
Participants consistently reported that the annotation process surfaced attentional patterns they had not previously recognized, generating new self-insight rather than simply recording pre-existing knowledge. Several described encountering gaze-shift tendencies for the first time---recognizing an unconscious habit of disengaging when scenes became predictable, or realizing their switching behavior followed patterns never consciously registered (P4, P10). Participants highlighted the value of the \textit{why-leave/why-arrive} structure for reconstructing shifts from both perspectives, and noted that classifying shifts as reactive or proactive further deepened reflection. Reviewing the AI-generated summary helped participants reconnect with thoughts from monitoring (P1, P9).

\textbf{Theme 2:} \textit{Retrospective Reconstruction.}
While participants could generally produce plausible explanations for gaze shifts, several questioned whether these faithfully reflected in-the-moment reasoning, noting that knowing the clip outcome made it difficult to disentangle genuine recall from post-hoc justification (P2, P7). P8 captured this ambiguity: \quotes{I'm not sure if it really helped me remember, or if it makes me fake remember---I cannot tell the difference.} The replay function and gaze-shift timeline partially addressed this by restoring attentional context at the point of annotation (P5, P9), while confidence and memory clarity ratings helped participants avoid overclaiming certainty. 

\textbf{Theme 3:} \textit{Robot Intentionality as Gaze Anchor.}
To identify the primary drivers of gaze shifts, participants ranked four contextual cues by importance. Using Borda count aggregation ($N = 12$)~\cite{emerson2013original}, \textit{Robot Context} ranked highest (6/12 first-place rankings; score\,=\,28), followed by \textit{Visual Context} (4/12; score\,=\,24), \textit{Task Context} (2/12; score\,=\,20), and \textit{Others} (0/12; score\,=\,0), suggesting that robot behavior was the dominant attentional cue, with visual salience having secondary importance.
Gaze allocation was anchored to participants' real-time interpretation of robot intent: when behavior was legible, operators directed sustained attention; when intent became ambiguous, gaze lost its anchor and became passive. P11 noted feeling bored when a robot became stuck, but re-engaging the moment it appeared to discover a new object. P1 ranked \textit{Task Context} lowest because \quotes{most of the time I couldn't understand what the robot was trying to do.} These findings suggest that operator attention profiles are not stable personal traits, but are co-constructed with the legibility of robot behavior.

\textbf{Theme 4:} \textit{Accuracy Threshold for AI Annotation as Cognitive Scaffold.}
Participants responded positively to AI-generated summaries when the summaries compressed behavior into an interpretable profile. P3 described the result as feeling \textit{``like my profile,''} and P5 noted it \textit{``aligns perfectly with the summary the annotation generated for me.''} However, participants noted that current outputs were more descriptive than interpretive (P9) and sometimes conflated object- and scene-level attention (P10). A clear accuracy threshold emerged: P9 estimated roughly \textit{``80 or 90\% accuracy''} would be needed before preferring AI annotation over manual effort, and also noted a cognitive tradeoff---\textit{``although I take less time with AI annotation\ldots I feel more cognitive load.''} Participants favored a validation-oriented workflow (P6) over full automation, suggesting AI support should be framed around efficient review rather than replacement.

\subsubsection{Usability and Usefulness}
\system{} received a mean SUS score of 75.21 ($SD = 14.48$), corresponding approximately to \citet{bangor2008empirical}'s \textit{Good} anchor.
Several \system~annotation components were widely valued: the replay function and gaze-shift timeline helped restore attentional context at the point of annotation (P1, P5, P9), and the why-leave/why-arrive structure was preferred over single-question formats for its ability to
reconstruct a shift from both perspectives (P6, P8). The primary usability complaint was information density---P10 proposed progressive
disclosure via step-by-step pop-ups, noting that presenting all options at once confused new users.
Participants rarely struggled to describe gaze-shift reasoning in
natural language; difficulty emerged when mapping explanations onto predefined categories. P9 found fine-grained labels---such as
safety type, role, or task---difficult to distinguish, while P5 noted
that the full option set appeared regardless of clip context and argued that the system should dynamically narrow viable choices.

%% file: 5_discussion.tex
\section{Discussion}
{\bl \system{} is a pre-deployment elicitation tool that helps operators retrospectively reconstruct and characterize their gaze shifts during multi-robot monitoring.}
{\bl It is conceived as an initial step toward a broader calibration workflow in which designers may ultimately use operator-confirmed attention profiles to inform robot behavior and interface design before deployment.}
\puqi{§5 to frame Attune as a pre-deployment elicitation tool that surfaces and structures operators' retroactive accounts of gaze shifts into 
“attention profiles,” leaving downstream uses to future work.}
Our evaluation {\bl helps reveal} the viability of this vision and surfaces three design implications for calibration systems. We discuss these implications below.
\subsection{Implication: Annotation Elicits Rationales}
A central finding is that retrospective, calibration-oriented gaze-shift attribution does not merely \textit{record} operator behavior; it can also help \textit{elicit} operator knowledge. Participants often recognized attentional tendencies during annotation that they had not explicitly articulated before, such as disengaging when a scene became predictable or switching views in response to cues they had not consciously registered (P4, P10). This suggests that calibration may elicit tacit knowledge that is practice-based, context-dependent, and not readily accessible through direct introspection alone~\cite{polanyi2009tacit}.

\system{}'s annotation workflow appears to support this process in several ways. The \textit{why-leave/why-arrive} structure helped participants interpret gaze shifts from both the source and destination perspectives, while the reactive/proactive classification encouraged reflection on broader top-down and bottom-up patterns rather than isolated events. Together, these findings suggest that calibration-oriented annotation supports event-level explanation rather than merely recording where operators look.

At the same time, retrospective annotation introduces a methodological tension: participants may reconstruct plausible explanations rather than retrieve original in-the-moment reasoning, especially when clip outcomes are already known (P8). 
{\bl This post-hoc approach was nevertheless deliberate, as requesting explanations during monitoring would interrupt the monitoring task and could alter the gaze behavior being observed. One alternative that avoids directly interrupting the monitoring task is passive inference over substantially more data than a short calibration session can provide. Retrospective annotation therefore trades a risk of recall decay for gaze behavior that remains uncontaminated by the act of measurement.} 
\puqi{Addressing R2/2AC’s concerns, we will justify post-hoc annotation in §4/§\textbf{5.1} as a design choice to avoid interrupting monitoring or altering natural gaze behavior.}
Our findings suggest that replay, nonlinear timeline magnification, and confidence or memory-clarity ratings may help mitigate this risk, but do not eliminate it. Video length also remained a consistent source of recall degradation. More broadly, these results suggest that calibration should be treated not as passive data collection, but as a structured elicitation process that helps operators externalize otherwise implicit monitoring strategies while accounting for the limits of retrospective recall.
\puqi{ §5 to clarify Attune’s scope. (2) We will clarify that Attune does not assume the monitoring duration after the calibration step; its elicited insights are intended to apply to both longer and shorter bursts of sustained monitoring (e.g., immediately post-alert) (3) We will clarify that Attune can complement alerts by helping understand how to sustain attention once it has been drawn and avoid operator distraction. Future work will discuss integrating with alert-driven workflows.
}

\textbf{Design Recommendation:} Calibration systems should explicitly support operator memory during retrospective annotation, for example, through visual signposting, rapid replay around shift boundaries, and explicit uncertainty marking for later review.

\subsection{Implication: Robot Legibility and Attention}
Our findings suggest that operator attention profiles should not be interpreted as purely internal preferences, but are co-shaped through interaction with robot behavior and how legibly it communicates state and intent. What operators choose to monitor may partially depend on how clearly the robot communicates its state, intention, and ongoing activity to the operator.

This matters because a gaze shift can reflect different causes. In some cases, it indicates purposeful monitoring of meaningful robot activity. In others, it reflects an effort to resolve uncertainty caused by ambiguous motion, weak signaling, or limited visual evidence. Seen this way, calibration does not simply capture where an operator prefers to look; it can also reveal where robot behavior or multi-view presentation fails to support efficient supervision.

More broadly, these findings suggest that calibration should be treated as a relational process rather than a one-sided profiling method. An attention profile may reflect the fit between operator expectations and robot legibility. Repeated shifts toward moments of ambiguity, verification, or missed cues can indicate not only attentional tendencies, but also opportunities to redesign robot behavior for more interpretable supervision.

\textbf{Design Recommendation:} Calibration systems should account for robot behavioral legibility when interpreting attention profiles, so that attention patterns are understood as a function of both operator strategy and the clarity of robot intent.

\subsection{Implication: AI Assistance Scaffolds and Shapes Reflection}
Our findings suggest that AI assistance can make calibration more scalable by reducing annotation effort and helping operators recognize broader attention patterns that may be difficult to extract from raw annotations alone. Participants often found AI-generated summaries useful for making their monitoring tendencies more legible, suggesting that the pipeline can transform fragmented annotations into 
{\bl meaningful inputs for future design work.}
\puqi{leaving downstream uses to future work. }
At the same time, the value of AI assistance appears contingent on suggestion quality, where participants expressed that AI usefulness depended on whether a suggestion matched their own recollections. Suggestions that matched participants’ recollections appeared to reduce effort and support reflection, whereas mismatches increased perceived verification burden and could undermine the validity of the calibration itself. This risk was especially pronounced when participants no longer remembered the original gaze event clearly enough to confidently confirm or reject a suggestion. In such cases, AI assistance risks introducing plausible but unverifiable interpretations that could shape, rather than merely support, the operator's explanation of their own attentional reasoning.
{\bl This concern motivates bolstering human-in-the-loop calibration, in which inadequately calibrated models may produce plausible but spurious rationales. In our case, rather than inferring subjective intent from scratch, Phase 4 propagates rationales anchored in the operator’s manual annotations and reviewed patterns. This calibration incurs upfront effort; future mixed-initiative approaches could reduce that cost by prioritizing ambiguous or informative shifts for manual review.
}
\puqi{Regarding R2/1AC’s concerns about manual annotation practicality, we will highlight Attune as a one-time, human-in-the-loop calibration step to elicit operator-endorsed rationales to support future full automation. We will emphasize in §5.3 that uncalibrated AI risks producing spurious rationales when inferring subjective intent without human input. Instead, Attune auto-propagates human-anchored rationales in Phase 4.  We will acknowledge calibration cost and outline future mixed-initiative approaches to cut it.}

Overall, our findings suggest that AI-assisted calibration should be designed as a human-in-the-loop, verification-oriented workflow, thus scaffolding reflection, rather than a purely automated workflow. {\bl Automated gaze analysis answers where and when the eye moved, but the rationales participants surfaced were tacit and not present in any signal. Annotation and pattern analysis provides this signal, suggesting why an operator's gaze shifts. 
As such, we view calibration as necessary for telling a designer what to change.
We argue that the initial cost of acquiring this signal is outweighed by its benefit, as it grounds later automation in the operator's own confirmed rationales rather than in inferences from external cues.}
\puqi{Regarding R2/1AC’s concerns about manual annotation practicality, we will highlight Attune as a one-time, human-in-the-loop calibration step to elicit operator-endorsed rationales to support future full automation. We will emphasize in §5.3 that uncalibrated AI risks producing spurious rationales when inferring subjective intent without human input. Instead, Attune auto-propagates human-anchored rationales in Phase 4. }
After manual annotation, \system{} positions the operator as the final interpretive authority, thus preserving a verification-oriented workflow. To ease future verification burdens, low-confidence AI suggestions with low information gain can be suppressed. 

Finally, participants' privacy concerns suggest that this reflective support must also be bounded in its long-term use. Repeated calibration across deployments may enable increasingly fine-grained behavioral profiles of individual operators, raising concerns about secondary use beyond the original calibration context. Calibration systems should therefore make data retention limits, access boundaries, {\bl deletion after the calibration session, limits on sharing raw gaze video,} and intended uses explicit, {\bl particularly given the risk of repurposing calibration data for workplace surveillance,} so that the personalization that makes calibration valuable does not become a liability for the people it is meant to support.
\puqi{we will add a privacy discussion in §5.3 addressing data retention, access, deletion after calibration, and limits on sharing(2AC).}

\textbf{Design Recommendation:} AI-assisted calibration systems must prioritize and support human-in-the-loop verification over pure automation by surfacing high-confidence and high-information-gain suggestions, preserving operator authority over interpretation, and supporting operator privacy.
\subsection{Limitations and Future Work}
This work has several limitations.
{\bl While our two-robot, six-view, prerecorded setup represents the minimal multi-robot configuration needed to elicit cross-robot gaze shifts and enables a foundational feasibility test, it does not establish transfer to larger fleets, live deployments, different robot platforms, or alternative display layouts, all of which may change workload and gaze behavior. Future work should examine different fleet sizes, platforms, and display layouts.}
\puqi{large-fleet claims and frame the two-robot setup as a foundational first step for validating general feasibility, and the minimal configuration that produces the cross-robot gaze shifts necessary to evaluate Attune’s elicitation workflow. We will discuss scaling to larger fleets, different robot platforms, and varied monitor layouts as future work. }

Much additional analysis remains to be done, including evaluating the objective quality of summaries produced by \system{}. Crucially, we did not leverage our data beyond analysis, and as a result, our contributions stop short of demonstrating how attention profiles can be operationalized in practice. Important next steps are to (1) evaluate these profiles with interface designers, (2) inform the development of multi-robot video sensemaking tools tailored to individual operator attentional tendencies, and (3) adapt robot behaviors directly in response to these profiles. {\bl Future work should also integrate \system{} with alert-driven workflows and investigate how our proposed calibration step can sustain attention at the alert source while avoiding distraction elsewhere.}
\puqi{(3) We will clarify that Attune can complement alerts by helping understand how to sustain attention once it has been drawn and avoid operator distraction. Future work will discuss integrating with alert-driven workflows.}

Additional limitations remain. First, \system{} supports retrospective rather than live annotation. While retroactive annotation facilitates reflection, it also risks memory decay and inaccurate rationalization. Future work should explore live or near-real-time annotation despite its potential cognitive cost.
Second, we did not systematically evaluate LLM accuracy. Our findings therefore focus on participants' use of AI assistance rather than model performance itself. Future work should examine how LLM accuracy affects calibration validity, verification effort, and trust. {\bl Third, the study did not include task-based or visual-saliency baselines, so it cannot objectively distinguish task-driven from salience-driven shifts; future work should compare operator-endorsed rationales with such baselines.} 
{\bl Fourth, \system{} is a purpose-built research prototype rather than an extension of established behavioral annotation tooling. Building a custom interface was necessary due to the lack of existing tools that couple gaze-shift detection, robot replay, and annotation in a single workflow (\S2.3). It nevertheless raises an adoption cost: practitioners cannot access these capabilities from the tools they already use, and our usability findings are specific to our interface rather than to gaze-rationale annotation in general. Future work should integrate our approach into existing tools.}
{\bl Fifth, \system{} does not assume a specific monitoring duration after calibration---its elicited profiles are intended to apply both to sustained monitoring and shorter windows. Though shorter, alert-driven windows were not evaluated here, alerts and gaze calibration are a promising research direction. Alerts indicate where attention is needed, whereas \system{} characterizes how attention is sustained or lost once a feed is inspected. Integrating alert logs as a secondary timeline is therefore a natural extension. For longer monitoring durations, the manual cost of calibration could be reduced through further mixed-initiative approaches.}
Lastly, our study design has other limitations. Verbalization increases cognitive burden
{\bl and divergent verbalization strategies may have altered natural attention allocation. Future work should explore non-verbalized designs. Additionally, we recruited na\"ive participants, whereas domain experts are needed to assess applicability to operational multi-robot contexts. Larger per-scenario samples and additional measures beyond our current study are also needed to further strengthen our analyses and fully validate \system{}'s profiling capability.}
\puqi{NOTE THIS MOVED TO LIMITATION SECTION: We will clarify in §4 that additional measures are needed to fully validate Attune’s profiling capability.}
\puqi{Regarding R1's concern that verbalization confounds visual attention/individual variation, we will acknowledge this limitation in §5.4, walk back individual-variation claims in §4.4.2, and sharpen our justification for verbalization as maintaining on-screen engagement. Non-verbalized study designs will be designated as future work. We will reframe participants as naive viewers (R1), scope claims accordingly, and list a study with domain experts as future work.}
%

%% file: cite.bib
@inproceedings{lewis2014task,
  title={Task switching and single vs. multiple alarms for supervisory control of multiple robots},
  author={Lewis, Michael and Chien, Shi-Yi and Mehortra, Siddarth and Chakraborty, Nilanjan and Sycara, Katia},
  booktitle={International conference on engineering psychology and cognitive ergonomics},
  pages={499--510},
  year={2014},
  doi={10.1007/978-3-319-07515-0_50},
  publisher={Springer International Publishing}
}

@inproceedings{chien2011effects,
  title={Effects of alarms on control of robot teams},
  author={Chien, Shih-Yi and Wang, Huadong and Lewis, Michael and Mehrotra, Siddharth and Sycara, Katia},
  booktitle={Proceedings of the Human Factors and Ergonomics Society Annual Meeting},
  volume={55},
  number={1},
  pages={434--438},
  year={2011},
  doi={10.1177/1071181311551089},
  publisher={Sage Publications Sage CA: Los Angeles, CA}
}

@article{emerson2013original,
  title={The original Borda count and partial voting},
  author={Emerson, Peter},
  journal={Social Choice and Welfare},
  volume={40},
  number={2},
  pages={353--358},
  year={2013},
  doi={10.1007/s00355-011-0603-9 },
  publisher={Springer}
}

@inproceedings{pelikan2024designing,
  title={Designing human-robot interactions: a StEER tutorial},
  author={Pelikan, Hannah and Winkle, Katie and Porfirio, David},
  booktitle={Adjunct Proceedings of the 2024 Nordic Conference on Human-Computer Interaction},
  pages={1--4},
  doi={10.1145/3677045.3685467},
  year={2024}
}

@inproceedings{pelikan2024encountering,
  title={Encountering autonomous robots on public streets},
  author={Pelikan, Hannah RM and Reeves, Stuart and Cantarutti, Marina N},
  booktitle={Proceedings of the 2024 ACM/IEEE International Conference on Human-Robot Interaction},
  pages={561--571},
  doi={10.1145/3610977.3634936},
  year={2024}
}

@inproceedings{passero2024honkable,
  title={Honkable Gestalts: Why autonomous vehicles get honked at},
  author={Passero, Sergio and Pelikan, Hannah Rm and Broth, Mathias and Brown, Barry},
  booktitle={Proceedings of the 16th International Conference on Automotive User Interfaces and Interactive Vehicular Applications},
  pages={317--328},
  doi={10.1145/3640792.3675732 },
  year={2024}
}

@inproceedings{zhang2025rosannotator,
  title={ROSAnnotator: A Web Application for ROSBag Data Analysis in Human-Robot Interaction},
  author={Zhang, Yan and Li, Haoqi and Tabatabaei, Ramtin and Johal, Wafa},
  booktitle={2025 20th ACM/IEEE International Conference on Human-Robot Interaction (HRI)},
  pages={1099--1103},
  year={2025},
  doi={10.1109/hri61500.2025.10974254},
  organization={IEEE}
}

@software{atlasti,
  author = {{ATLAS.ti}},
  title = {ATLAS.ti Scientific Software Development GmbH},
  url = {https://atlasti.com},
  version = {23.2.1},
  date = {2023},
}

@misc{elan,
  author       = {{Max Planck Institute for Psycholinguistics}},
  title        = {{ELAN} (Version 7.1) [Computer software]},
  year         = {2026},
  address      = {Nijmegen, The Netherlands},
  publisher    = {The Language Archive},
  howpublished = {\url{https://archive.mpi.nl/tla/elan}},
}

@article{mcilvenny2024dotebase,
  title={DOTEbase: software tools for qualitative analysis},
  author={McIlvenny, Paul and Davidsen, Jacob Gorm and Stein, Alexander},
  year={2024}
}

@article{mcilvenny2024guide,
  title={Guide for DOTEbase users: An online help guide},
  author={McIlvenny, Paul},
  year={2024}
}

@article{mcilvenny2022dote,
  title={DOTE: distributed open transcription environment},
  author={McIlvenny, Paul and Davidsen, Jacob Gorm and Stein, Alexander and Kov{\'a}cs, Art{\'u}r Barnab{\'a}s},
  year={2022}
}

@inproceedings{sloetjes2008annotation,
  title={Annotation by category-ELAN and ISO DCR},
  author={Sloetjes, Han and Wittenburg, Peter},
  booktitle={6th international Conference on Language Resources and Evaluation (LREC 2008)},
  year={2008},
  doi={10.63317/454o8uo6f7z8}
}

@article{friard2016boris,
  title={BORIS: a free, versatile open-source event-logging software for video/audio coding and live observations},
  author={Friard, Olivier and Gamba, Marco},
  journal={Methods in ecology and evolution},
  volume={7},
  number={11},
  pages={1325--1330},
  year={2016},
  doi={10.1111/2041-210x.12584 },
  publisher={Wiley Online Library}
}

@article{dhakal2022nvivo,
  title={NVivo},
  author={Dhakal, Kerry},
  journal={Journal of the Medical Library Association: JMLA},
  volume={110},
  number={2},
  pages={270},
  doi={10.5195/jmla.2022.1271},
  year={2022}
}

@article{lewis2007qda,
  title={QDA Miner 2.0: Mixed-model qualitative data analysis software},
  author={Lewis, R Barry and Maas, Steven M},
  journal={Field methods},
  volume={19},
  number={1},
  pages={87--108},
  year={2007},
  doi={10.1177/1525822x06296589 },
  publisher={Sage Publications Sage CA: Thousand Oaks, CA}
}

@article{barz2025eyenotate,
  title={eyeNotate: Interactive Annotation of Mobile Eye Tracking Data Based on Few-Shot Image Classification},
  author={Barz, Michael and Bhatti, Omair Shahzad and Alam, Hasan Md Tusfiqur and Nguyen, Duy Minh Ho and Altmeyer, Kristin and Malone, Sarah and Sonntag, Daniel},
  journal={Journal of Eye Movement Research},
  volume={18},
  number={4},
  pages={27},
  year={2025},
  doi={10.3390/jemr18040027 },
  publisher={MDPI}
}

@article{niehorster2025gazeMapper,
    Author = {Niehorster, Diederick C. and
              Hessels, R. S. and
              Nystr{\"o}m, Marcus and
              Benjamins, J. S. and
              Hooge, I. T. C.},
    Journal = {Behavior Research Methods},
    Number = {},
    Title = {{gazeMapper}: A tool for automated world-based analysis of gaze data from one or multiple wearable eye trackers},
    Year = {2025},
    doi = {10.3758/s13428-025-02704-4}
}

@inproceedings{somashekarappa2020annotation,
  title={An annotation approach for social and referential gaze in dialogue},
  author={Somashekarappa, Vidya and Howes, Christine and Sayeed, Asad},
  booktitle={Proceedings of the Twelfth Language Resources and Evaluation Conference},
  pages={759--765},
  year={2020}
}

@article{deane2023deep,
  title={Deep-SAGA: a deep-learning-based system for automatic gaze annotation from eye-tracking data},
  author={Deane, Oliver and Toth, Eszter and Yeo, Sang-Hoon},
  journal={Behavior Research Methods},
  volume={55},
  number={3},
  pages={1372--1391},
  year={2023},
  doi={10.3758/s13428-022-01833-4 },
  publisher={Springer}
}

@article{iddrisu2026eye,
  title={Eye Movement Classification Using Neuromorphic Vision Sensors},
  author={Iddrisu, Khadija and Shariff, Waseem and Stec, Maciej and O’Connor, Noel and Little, Suzanne},
  journal={Journal of Eye Movement Research},
  volume={19},
  number={1},
  pages={17},
  year={2026},
  doi={10.3390/jemr19010017 },
  publisher={Multidisciplinary Digital Publishing Institute}
}

@article{vortmann2021imaging,
  title={Imaging time series of eye tracking data to classify attentional states},
  author={Vortmann, Lisa-Marie and Knychalla, Jannes and Annerer-Walcher, Sonja and Benedek, Mathias and Putze, Felix},
  journal={Frontiers in Neuroscience},
  volume={15},
  pages={664490},
  year={2021},
  doi={10.3389/fnins.2021.664490 },
  publisher={Frontiers Media SA}
}

@inproceedings{pelikan2025people,
  title={The people behind the robots: How wizards wrangle robots in public deployments},
  author={Pelikan, Hannah RM and Bu, Fanjun and Ju, Wendy},
  booktitle={Proceedings of the 2025 CHI Conference on Human Factors in Computing Systems},
  pages={1--21},
  doi={10.1145/3706598.3713237 },
  year={2025}
}

@inproceedings{lee2025minding,
  title={Minding the Stop-Gap: Attending to the “Temporary,” Unplanned, and Added Labor of Human-Robot Collaboration in Context},
  author={Lee, Hee Rin and Fox, Sarah and Cheon, EunJeong and Shorey, Samantha},
  booktitle={2025 20th ACM/IEEE International Conference on Human-Robot Interaction (HRI)},
  pages={34--44},
  year={2025},
  doi={10.1109/hri61500.2025.10973955 },
  organization={IEEE}
}

@inproceedings{benford2025charting,
  title={Charting the Ecosystem of Trust in Cat Royale Or What It Takes to Trust a Robot to Play with Cats},
  author={Benford, Steve and Barnard, Pepita and Sharples, Sarah and Webb, Helena and Mancini, Clara and Kucukyilmaz, Ayse and Green, Simon Castle and Schneiders, Eike and Ngo, Victor and Chamberlain, Alan and others},
  booktitle={International Conference on Social Robotics},
  pages={623--636},
  year={2025},
  doi={10.1007/978-981-95-2398-6_42},
  organization={Springer}
}

@article{ebadi2023present,
  title={Present and future of SLAM in extreme environments: The DARPA SubT challenge},
  author={Ebadi, Kamak and Bernreiter, Lukas and Biggie, Harel and Catt, Gavin and Chang, Yun and Chatterjee, Arghya and Denniston, Christopher E and Desch{\^e}nes, Simon-Pierre and Harlow, Kyle and Khattak, Shehryar and others},
  journal={IEEE Transactions on Robotics},
  volume={40},
  pages={936--959},
  year={2023},
  doi={10.1109/tro.2023.3323938 },
  publisher={IEEE}
}

@inproceedings{gamboa2025we,
  title={We Are the Robots: Tapping Into the Lived Experiences of Wizards of Oz},
  author={Gamboa, Mafalda and Thunberg, Sofia and Alves-Oliveira, Patricia and Loerakker, Meagan B},
  booktitle={Proceedings of the Extended Abstracts of the CHI Conference on Human Factors in Computing Systems},
  pages={1--8},
  doi={10.1145/3706599.3720149 },
  year={2025}
}

@article{elbeleidy2023beyond,
  title={Beyond the session: Centering teleoperators in socially assistive robot-child interactions reveals the bigger picture},
  author={Elbeleidy, Saad and Mott, Terran and Liu, Dan and Do, Ellen and Reddy, Elizabeth and Williams, Tom},
  journal={Proceedings of the ACM on Human-Computer Interaction},
  volume={7},
  number={CSCW2},
  pages={1--33},
  year={2023},
  doi={10.1145/3610175 },
  publisher={ACM New York, NY, USA}
}

@incollection{polanyi2009tacit,
  title={The tacit dimension},
  author={Polanyi, Michael},
  booktitle={Knowledge in organisations},
  pages={135--146},
  year={2009},
  publisher={Routledge}
}

@article{bangor2008empirical,
  title={An empirical evaluation of the system usability scale},
  author={Bangor, Aaron and Kortum, Philip T and Miller, James T},
  journal={Intl. Journal of Human--Computer Interaction},
  volume={24},
  number={6},
  pages={574--594},
  year={2008},
  doi={10.1080/10447310802205776},
  publisher={Taylor \& Francis}
}

@article{braun2019thematic,
  title={Thematic analysis 48},
  author={Braun, Virginia and Clarke, Victoria and Hayfield, Nikki and Terry, Gareth},
  journal={Handbook of research methods in health social sciences},
  pages={843--860},
  year={2019},
  doi={10.1007/978-981-10-5251-4_103},
  publisher={Springer Singapore}
}

@inproceedings{walker2024cyber,
  title={The cyber-physical control room: A mixed reality interface for mobile robot teleoperation and human-robot teaming},
  author={Walker, Michael E and Gramopadhye, Maitrey and Ikeda, Bryce and Burns, Jack and Szafir, Daniel},
  booktitle={Proceedings of the 2024 ACM/IEEE International Conference on Human-Robot Interaction},
  pages={762--771},
  doi={10.1145/3610977.3634981 },
  year={2024}
}

@inproceedings{pelikan2025making,
  title={Making sense of public space for robot design},
  author={Pelikan, Hannah RM and Mutlu, Bilge and Reeves, Stuart},
  booktitle={2025 20th ACM/IEEE International Conference on Human-Robot Interaction (HRI)},
  pages={152--162},
  year={2025},
  organization={IEEE},
  doi={10.1109/hri61500.2025.10973847 }
}

@article{benford2025tangles,
  title={Tangles: Unpacking extended collision experiences with soma trajectories},
  author={Benford, Steve and Garrett, Rachael and Li, Christine and Tennent, Paul and N{\'u}nez-Pacheco, Claudia and Kucukyilmaz, Ayse and Tsaknaki, Vasiliki and H{\"o}{\"o}k, Kristina and Caleb-Solly, Praminda and Marshall, Joe and others},
  journal={ACM Transactions on Computer-Human Interaction},
  volume={32},
  number={4},
  pages={1--34},
  year={2025},
  doi={10.1145/3723875 },
  publisher={ACM New York, NY}
}

@inproceedings{kaya2023design,
  title={Design of an eight-wheeled mobile delivery robot and its climbing simulations},
  author={Kaya, Omer Mutlu Turk and Erdemir, Gokhan},
  booktitle={SoutheastCon 2023},
  pages={895--900},
  year={2023},
  doi={10.1109/southeastcon51012.2023.10115114 },
  organization={IEEE}
}

@article{valdez2023humans,
  title={Humans, robots and artificial intelligences reconfiguring urban life in a crisis},
  author={Valdez, Miguel and Cook, Matthew},
  journal={Frontiers in Sustainable Cities},
  volume={5},
  pages={1081821},
  year={2023},
  doi={10.3389/frsc.2023.1081821 },
  publisher={Frontiers Media SA}
}

@article{grimm2021practicalities,
  title={On the practicalities of robots in public spaces},
  author={Grimm, Cindy M and Thomasen, Kristen},
  journal={WeRobot 2021},
  year={2021}
}

@article{peters2015human,
  title={Human supervisory control of robotic teams: Integrating cognitive modeling with engineering design},
  author={Peters, Jeffrey R and Srivastava, Vaibhav and Taylor, Grant S and Surana, Amit and Eckstein, Miguel P and Bullo, Francesco},
  journal={IEEE Control Systems Magazine},
  volume={35},
  number={6},
  pages={57--80},
  year={2015},
  doi={10.1109/mcs.2015.2471056 },
  publisher={IEEE}
}

@inproceedings{ahlskog2024fostering,
  title={Fostering trust through user interface design in multi-drone search and rescue},
  author={Ahlskog, Johanna and Bahodi, Maria-Theresa and Lugmayr, Artur and Merritt, Timothy},
  booktitle={Proceedings of the Second International Symposium on Trustworthy Autonomous Systems},
  pages={1--11},
  doi={10.1145/3686038.3686052 },
  year={2024}
}

@article{chiou2021mixed,
  title={Mixed-initiative variable autonomy for remotely operated mobile robots},
  author={Chiou, Manolis and Hawes, Nick and Stolkin, Rustam},
  journal={ACM Transactions on Human-Robot Interaction (THRI)},
  volume={10},
  number={4},
  pages={1--34},
  year={2021},
  doi={10.1145/3472206 },
  publisher={ACM New York, NY, USA}
}

@inproceedings{drury2007adapting,
  title={Adapting GOMS to model human-robot interaction},
  author={Drury, Jill L and Scholtz, Jean and Kieras, David},
  booktitle={Proceedings of the ACM/IEEE international conference on Human-robot interaction},
  pages={41--48},
  doi={10.1145/1228716.1228723 },
  year={2007}
}

@article{drury2007modeling,
  title={Modeling Human-Robot Interaction with GOMS},
  author={Drury, Jill L and Scholtz, Jean and Kieras, David},
  year={2007}
}

@inproceedings{roy2023need,
  title={I need your help... or do i? maintaining situation awareness through performative autonomy},
  author={Roy, Sayanti and Smith, Trey and Coltin, Brian and Williams, Tom},
  booktitle={Proceedings of the 2023 ACM/IEEE international conference on human-robot interaction},
  pages={122--131},
  doi={10.1145/3568162.3576954 },
  year={2023}
}

@inproceedings{zhou2026designing,
  title={Designing Multi-Robot Ground Video Sensemaking with Public Safety Professionals},
  author={Zhou, Puqi and Asgarov, Ali and Hussain, Aafiya and Park, Wonjoon and Paudyal, Amit and Shrestha, Sameep and Tang, Chia-Wei and Lighthiser, Michael and Hieb, Michael and Xiao, Xuesu and others},
  booktitle={Proceedings of the 2026 CHI Conference on Human Factors in Computing Systems},
  pages={1--22},
  doi={10.1145/3772318.3790679 },
  year={2026}
}

@article{chiou2022towards,
  title={Towards human--robot teaming: Tradeoffs of explanation-based communication strategies in a virtual search and rescue task},
  author={Chiou, Erin K and Demir, Mustafa and Buchanan, Verica and Corral, Christopher C and Endsley, Mica R and Lematta, Glenn J and Cooke, Nancy J and McNeese, Nathan J},
  journal={International Journal of Social Robotics},
  volume={14},
  number={5},
  pages={1117--1136},
  year={2022},
  doi={10.1007/s12369-021-00834-1 },
  publisher={Springer}
}

@article{foroughi2023near,
  title={Near-perfect automation: Investigating performance, trust, and visual attention allocation},
  author={Foroughi, Cyrus K and Devlin, Shannon and Pak, Richard and Brown, Noelle L and Sibley, Ciara and Coyne, Joseph T},
  journal={Human factors},
  volume={65},
  number={4},
  pages={546--561},
  year={2023},
  doi={10.1177/00187208211032889 },
  publisher={Sage Publications Sage CA: Los Angeles, CA}
}

@article{wickens2021attention,
  title={Attention: Theory, principles, models and applications},
  author={Wickens, Christopher},
  journal={International Journal of Human--Computer Interaction},
  volume={37},
  number={5},
  pages={403--417},
  year={2021},
  doi={10.1080/10447318.2021.1874741 },
  publisher={Taylor \& Francis}
}

@article{rizzolatti1987reorienting,
  title={Reorienting attention across the horizontal and vertical meridians: evidence in favor of a premotor theory of attention},
  author={Rizzolatti, Giacomo and Riggio, Lucia and Dascola, Isabella and Umilt{\'a}, Carlo},
  journal={Neuropsychologia},
  volume={25},
  number={1},
  pages={31--40},
  year={1987},
  doi={10.1016/0028-3932(87)90041-8 },
  publisher={Elsevier}
}

@incollection{wickens2017attentional,
  title={Attentional models of multitask pilot performance using advanced display technology},
  author={Wickens, Christopher D and Goh, Juliana and Helleberg, John and Horrey, William J and Talleur, Donald A},
  booktitle={Human Error in Aviation},
  pages={155--175},
  year={2017},
  doi={10.4324/9781315092898-10 },
  publisher={Routledge}
}

@inbook{wickens2015noticing,
  title={Noticing events in the visual workplace: The SEEV and NSEEV models},
  author={Wickens, Christopher D},
  year={2015},
  doi={10.1017/cbo9780511973017.046},
  publisher={Cambridge University Press}
}

@article{wolfe2007guided,
  title={Guided search 4.0},
  author={Wolfe, Jeremy M and Gray, W},
  journal={Integrated models of cognitive systems},
  pages={99--119},
  year={2007}
}

@inproceedings{pei2025attentionar,
  title={AttentionAR: AR Adaptation and Warning for Real-World Safety via Attention Modeling and MLLM Reasoning},
  author={Pei, Yunqiang and Huang, Renming and Zha, Mingfeng and Wang, Guoqing and Wang, Peng and Kang, Qiao and Yang, Yang and Shen, Heng Tao},
  booktitle={Proceedings of the 38th Annual ACM Symposium on User Interface Software and Technology},
  pages={1--19},
  doi={10.1145/3746059.3747674 },
  year={2025}
}

@inproceedings{saad2019welcoming,
  title={Welcoming robot behaviors for drawing attention},
  author={Saad, Elie and Neerincx, Mark A and Hindriks, Koen V},
  booktitle={2019 14th ACM/IEEE International Conference on Human-Robot Interaction (HRI)},
  pages={636--637},
  year={2019},
  doi={10.1109/hri.2019.8673325 },
  organization={IEEE}
}

@inproceedings{miyashita2025framelight,
  title={FrameLight: Guiding Attention Through Motion and Arrangement in Shape-Changing Furniture Robots},
  author={Miyashita, Hyu and Nakanishi, Yasuto},
  booktitle={2025 20th ACM/IEEE International Conference on Human-Robot Interaction (HRI)},
  pages={1498--1502},
  year={2025},
  doi={10.1109/hri61500.2025.10974138 },
  organization={IEEE}
}

@inproceedings{yamaoka2009developing,
  title={Developing a model of robot behavior to identify and appropriately respond to implicit attention-shifting},
  author={Yamaoka, Fumitaka and Kanda, Takayuki and Ishiguro, Hiroshi and Hagita, Norihiro},
  booktitle={Proceedings of the 4th ACM/IEEE international conference on Human robot interaction},
  pages={133--140},
  doi={10.1145/1514095.1514120},
  year={2009}
}

@inproceedings{lemaignan2016real,
  title={From real-time attention assessment to “with-me-ness” in human-robot interaction},
  author={Lemaignan, S{\'e}verin and Garcia, Fernando and Jacq, Alexis and Dillenbourg, Pierre},
  booktitle={2016 11th ACM/IEEE international conference on human-robot interaction (HRI)},
  pages={157--164},
  year={2016},
  doi={10.1109/hri.2016.7451747 },
  organization={Ieee}
}

@inproceedings{schirmer2025utilizing,
  title={Utilizing eye gaze for human-robot collaborative assembly},
  author={Schirmer, Fabian and Kranz, Philipp and Rose, Chad G and Willert, Volker and Schmitt, Jan and Kaupp, Tobias},
  booktitle={2025 20th ACM/IEEE International Conference on Human-Robot Interaction (HRI)},
  pages={1603--1607},
  year={2025},
  doi={10.1109/hri61500.2025.10974041 },
  organization={IEEE}
}

@inproceedings{rea2017movers,
  title={Movers, shakers, and those who stand still: visual attention-grabbing techniques in robot teleoperation},
  author={Rea, Daniel J and Seo, Stela H and Bruce, Neil and Young, James E},
  booktitle={Proceedings of the 2017 ACM/IEEE International Conference on Human-Robot Interaction},
  pages={398--407},
  doi={10.1145/2909824.3020246 },
  year={2017}
}

@inproceedings{ozsu2025distraction,
  title={Distraction by a Human or a Robot: Effects of Perceptual Load and Action Type},
  author={{\"O}zsu, Ataol Burak and Pek{\c{c}}etin, Tu{\u{g}}{\c{c}}e Nur and Faydali, Defne {\c{S}}iir and Urgen, Burcu A},
  booktitle={2025 20th ACM/IEEE International Conference on Human-Robot Interaction (HRI)},
  pages={1171--1175},
  year={2025},
  doi={10.1109/hri61500.2025.10974021 },
  organization={IEEE}
}

@article{sam2024impact,
  title={The impact of stress and workload on human performance in robot teleoperation tasks},
  author={Sam, Yi Ting and Hedlund-Botti, Erin and Natarajan, Manisha and Heard, Jamison and Gombolay, Matthew},
  journal={IEEE Transactions on Robotics},
  volume={40},
  pages={4725--4744},
  year={2024},
  doi={10.1109/tro.2024.3484630 },
  publisher={IEEE}
}

@incollection{diana2024designing,
  title={Designing robots that work and matter},
  author={Diana, Carla},
  booktitle={Designing Interactions with Robots},
  pages={140--147},
  year={2024},
  doi={10.1201/9781003371021-7 },
  publisher={Chapman and Hall/CRC}
}

@article{zhang2025new,
  title={A New Trend In the Warehousing: A Review of Human Robot Collaboration},
  author={Zhang, Yu and Xue, Songdong},
  journal={IEEE Access},
  year={2025},
  doi={10.1109/access.2025.3606967 },
  publisher={IEEE}
}

@article{wang2025concurrent,
  title={Concurrent Multi-Robot Search of Multiple Missing Persons in Urban Environments},
  author={Wang, Zicheng and Benhabib, Beno},
  journal={Robotics},
  volume={14},
  number={11},
  pages={157},
  year={2025},
  doi={10.3390/robotics14110157 },
  publisher={MDPI}
}

@article{crandall2010computing,
  title={Computing the effects of operator attention allocation in human control of multiple robots},
  author={Crandall, Jacob W and Cummings, Mary L and Della Penna, Mauro and De Jong, Paul MA},
  journal={IEEE Transactions on Systems, Man, and Cybernetics-Part A: Systems and Humans},
  volume={41},
  number={3},
  pages={385--397},
  year={2010},
  doi={10.1109/tsmca.2010.2084082 },
  publisher={IEEE}
}

@article{srivastava2014attention,
  title={Attention allocation for decision making queues},
  author={Srivastava, Vaibhav and Carli, Ruggero and Langbort, C{\'e}dric and Bullo, Francesco},
  journal={Automatica},
  volume={50},
  number={2},
  pages={378--388},
  year={2014},
  doi={10.1016/j.automatica.2013.11.028 },
  publisher={Elsevier}
}

@inproceedings{redmon2016you,
  title={You only look once: Unified, real-time object detection},
  author={Redmon, Joseph and Divvala, Santosh and Girshick, Ross and Farhadi, Ali},
  booktitle={Proceedings of the IEEE conference on computer vision and pattern recognition},
  pages={779--788},
  doi={10.1109/cvpr.2016.91 },
  year={2016}
}

@inproceedings{holler2020hddl,
  title={HDDL: An extension to PDDL for expressing hierarchical planning problems},
  author={H{\"o}ller, Daniel and Behnke, Gregor and Bercher, Pascal and Biundo, Susanne and Fiorino, Humbert and Pellier, Damien and Alford, Ron},
  booktitle={Proceedings of the AAAI conference on artificial intelligence},
  volume={34},
  number={06},
  pages={9883--9891},
  doi={10.1609/aaai.v34i06.6542 },
  year={2020}
}

@article{fox2003pddl2,
  title={PDDL2. 1: An extension to PDDL for expressing temporal planning domains},
  author={Fox, Maria and Long, Derek},
  journal={Journal of artificial intelligence research},
  volume={20},
  pages={61--124},
  doi={10.1613/jair.1129 },
  year={2003}
}

@inproceedings{salvucci2000identifying,
author = {Salvucci, Dario D. and Goldberg, Joseph H.},
title = {Identifying fixations and saccades in eye-tracking protocols},
year = {2000},
isbn = {1581132808},
publisher = {Association for Computing Machinery},
address = {New York, NY, USA},
url = {https://doi.org/10.1145/355017.355028},
doi = {10.1145/355017.355028},
booktitle = {Proceedings of the 2000 Symposium on Eye Tracking Research \& Applications},
pages = {71–78},
numpages = {8},
location = {Palm Beach Gardens, Florida, USA},
series = {ETRA '00}
}

@article{manor2003defining,
  title={Defining the temporal threshold for ocular fixation in free-viewing visuocognitive tasks},
  author={Manor, Barry R and Gordon, Evian},
  journal={Journal of neuroscience methods},
  volume={128},
  number={1-2},
  pages={85--93},
  year={2003},
  doi={10.1016/s0165-0270(03)00151-1 },
  publisher={Elsevier}
}

@inproceedings{wu2024theia,
  title={Theia: Gaze-driven and perception-aware volumetric content delivery for mixed reality headsets},
  author={Wu, Nan and Liu, Kaiyan and Cheng, Ruizhi and Han, Bo and Zhou, Puqi},
  booktitle={Proceedings of the 22nd Annual International Conference on Mobile Systems, Applications and Services},
  pages={70--84},
  doi={10.1145/3643832.3661858 },
  year={2024}
}

@ARTICLE{gao2014grid,
  author={Gao, Fei and Cummings, Mary L. and Solovey, Erin Treacy},
  journal={IEEE Transactions on Human-Machine Systems}, 
  title={Modeling Teamwork in Supervisory Control of Multiple Robots}, 
  year={2014},
  volume={44},
  number={4},
  pages={441-453},
  doi={10.1109/THMS.2014.2312391}}

@article{brooke1996sus,
  title={SUS-A `quick and dirty' usability scale},
  author={Brooke, John},
  journal={Usability Evaluation in Industry},
  volume={189},
  number={194},
  pages={4--7},
  year={1996},
  publisher={London, England},
  doi={10.1201/9781498710411-35 }
}

@INPROCEEDINGS{kemp2026stretch,
  author={Kemp, Charles C. and Edsinger, Aaron and Clever, Henry M. and Matulevich, Blaine},
  booktitle={2022 International Conference on Robotics and Automation (ICRA)}, 
  title={The Design of Stretch: A Compact, Lightweight Mobile Manipulator for Indoor Human Environments}, 
  year={2022},
  volume={},
  number={},
  pages={3150-3157},
  doi={10.1109/ICRA46639.2022.9811922}}
